\documentclass[twoside,journal]{IEEEtran}

\ifCLASSOPTIONcompsoc
    \usepackage[caption=false,font=normalsize,labelfont=sf,textfont=sf]{subfig}
\else
    \usepackage[caption=false,font=footnotesize]{subfig}
\fi

\usepackage[utf8]{inputenc}

\usepackage{amsmath,amsfonts,amssymb,bm,xspace,nicefrac}
\usepackage{amsthm}
\usepackage{cases}

\usepackage{microtype}
\usepackage{url}
\usepackage[hidelinks]{hyperref}
\usepackage{cite}

\usepackage{xcolor}
\usepackage{color}

\usepackage{enumitem}
\usepackage{booktabs}
\usepackage{colortbl}
\usepackage{multirow}
\usepackage{threeparttable}
\usepackage{graphicx}
\usepackage{draftfigure}
\usepackage{wrapfig}

\usepackage{stfloats}

\usepackage[switch]{lineno}

\usepackage{algorithm}
\usepackage{algpseudocode}

\newcommand{\Blue}[1]{{\color{blue}#1}}

\newcommand{\etal}{\textit{et al.}}
\newcommand{\eg}{\textit{e.g.}}
\newcommand{\ie}{\textit{i.e.}}

\theoremstyle{definition}
\newtheorem{definition}{Definition}

\begin{document}
\thispagestyle{empty}
\begin{figure*}[t!]
    \large{This work has been submitted to the IEEE for possible publication. Copyright may be transferred without notice, after which this version may no longer be accessible.}
\end{figure*}
\clearpage


\title{Knowledge Amalgamation for Object Detection with Transformers}
\author{
Haofei~Zhang,
Feng~Mao,
Mengqi~Xue,
Gongfan~Fang,
Zunlei~Feng,
Jie~Song,
Mingli~Song%
\thanks{%
Haofei~Zhang, Mengqi~Xue, Gongfan~Fang and Mingli Song, are with the College of Computer Science, Zhejiang university, Hangzhou, China (e-mail: \{haofeizhang, mqxue, fgfvain97, brooksong\}@zju.edu.cn).%
}%
\thanks{%
Feng~Mao is with the Alibaba Xixi Campus, Hangzhou, China (e-mail: maofeng.mf@alibaba-inc.com).%
}%
\thanks{%
Zunlei Feng is with the College of Software Technology, Zhejiang University, Hangzhou, China (e-mail: zunleifeng@zju.edu.cn)%
}%
\thanks{%
Jie~Song is with the College of Software Technology, Zhejiang University, Hangzhou, China and Zhejiang Lab, Hangzhou, China (e-mail: sjie@zju.edu.cn).%
}%
}

\markboth{Journal of \LaTeX\ Class Files,~Vol.~xx, No.~x, August~xxxx}{Zhang \MakeLowercase{\etal}}

\maketitle

\begin{abstract}
Knowledge amalgamation~(KA) is a novel deep model reusing task aiming to transfer knowledge from several well-trained teachers to a multi-talented and compact student.
Currently, most of these approaches are tailored for convolutional neural networks~(CNNs).
However, there is a tendency that transformers, with a completely different architecture, are starting to challenge the domination of CNNs in many computer vision tasks.
Nevertheless, directly applying the previous KA methods to transformers leads to severe performance degradation.
In this work, we explore a more effective KA scheme for transformer-based object detection models.
Specifically, considering the architecture characteristics of transformers, we propose to dissolve the KA into two aspects: sequence-level amalgamation~(SA) and task-level amalgamation~(TA).
In particular, a hint is generated within the sequence-level amalgamation by concatenating teacher sequences instead of redundantly aggregating them to a fixed-size one as previous KA works.
Besides, the student learns heterogeneous detection tasks through soft targets with efficiency in the task-level amalgamation.
Extensive experiments on PASCAL VOC and COCO have unfolded that the sequence-level amalgamation significantly boosts the performance of students, while the previous methods impair the students.
Moreover, the transformer-based students excel in learning amalgamated knowledge, as they have mastered heterogeneous detection tasks rapidly and achieved superior or at least comparable performance to those of the teachers in their specializations.

\end{abstract}

\begin{IEEEkeywords}
Model reusing, knowledge amalgamation, knowledge distillation, object detection, vision transformers.
\end{IEEEkeywords}

\ifCLASSOPTIONpeerreview
\begin{center} \bfseries EDICS Category: ARS-IVA \end{center}
\fi
\IEEEpeerreviewmaketitle

\section{Introduction}
\IEEEPARstart{W}{ith} the rapid development of deep learning~\cite{lecun2015deep}, a growing number of deep models have been trained and shared on the Internet. Reusing these pre-trained models to get slimmer ones for reducing computation cost has been a trending research topic in recent years.
Hinton \etal~\cite{hinton2015distilling} initially propose \textit{knowledge distillation}~(KD) for model compression, where the knowledge of a well-trained cumbersome teacher model is transferred to the lightweight student through soft target supervision provided by the teacher. Despite this simple approach, KD generally improves the performance of students with various architectures than training from scratch.
Following this teacher-student paradigm, FitNets~\cite{romero2014fitnets} utilize intermediate features, also known as \textit{hints}, to sufficiently supervise hidden layers of the student. Besides, a number of methods~\cite{zagoruyko2016paying,tian2019contrastive,song2021tree} have managed to extract deeper level of information from the intermediate layers and demonstrated promising results. Except for image classification, researchers are exploring KD towards many other tasks, such as semantic segmentation~\cite{liu2019structured,he2019knowledge}; object detection~\cite{chen2017learning,li2017mimicking,zhang2020improve,dai2021general}; natural language processing~\cite{jiao2019tinybert,sanh2019distilbert,sun2019patient}; and reinforcement learning~\cite{rusu2015policy,ashok2017n2n}.

Furthermore, there has been an increasing interest in \textit{knowledge amalgamation}~(KA), an extension of KD, where knowledge of several teachers is transferred to one multi-talent student~\cite{shen2019customizing,shen2019amalgamating,luo2019knowledge,ye2019amalgamating,ye2019student,jing2021amalgamating}.
For example,~\cite{shen2019customizing,shen2019amalgamating,luo2019knowledge} focus on training a student with complementary knowledge from homogeneous tasks, \eg, a couple of classification problems.
However, these methods share a common strategy: the intermediate student features are required to mimic the aggregated hints (usually achieved by linear projection). Specifically, Shen \etal~\cite{shen2019amalgamating} propose an auto-encoder structure for aggregate hints from different teachers. Moreover, Luo \etal~\cite{luo2019knowledge} aim at projecting all these features into a joint embedding space where they are closed to each other with the same input.
However, most works are proposed in the realm of convolutional neural networks~(CNNs). 

Recently, the \textit{Transformers}~\cite{NIPS2017_attention}, with great success in natural language processing (NLP)~\cite{NIPS2017_attention,brown2020gpt,devlin2018bert}, is introduced to various computer vision~(CV) tasks. Unlike CNNs that focus on localized features, transformers, which exclusively rely on the global attention mechanism, are natively capable of modeling the relationship between vision tokens~\cite{zhu2020deformable}. Nowadays, there is a tendency that vision transformers are starting to challenge the domination of CNNs in many CV tasks, for instance, ViT~\cite{dosovitskiy2021vit} in image classification and DETR~\cite{carion2020detr} in object detection. However, directly applying previous KA methods to transformers will degrade the student performance since the feature projection brings extra noise and loss of information (as shown in Section~\ref{sec:seq_amg_analysis} and~\ref{exp:ablation-agg}).

In this paper, we strive to explore an efficient and effective KA scheme towards transformer-based object detection models.
Our objective is to transfer knowledge from several well-trained teachers specialized in heterogeneous object detection tasks to a compact and versatile student as much as possible. As an example, assumed that one teacher detects vehicles and the other one detects animals, the student is therefore expected to detect both vehicles and animals.
To be specific, DETR, a fully end-to-end object detection model, is employed as the fundamental architecture for both teachers and the student due to the following two reasons. (1) DETR can efficiently predict a set of objects without hand-crafted procedures, such as predefining anchors and non-maximal suppression. As a result, it dramatically improves the training efficiency for the student. (2) When pre-trained as UP-DETR~\cite{dai2020up-detr}, DETR has the ability to achieve higher performance than traditional CNNs with fewer parameters.


Towards this end, we propose a novel knowledge amalgamation scheme with two individual components: sequence-level amalgamation~(SA) and task-level amalgamation~(TA).

\textit{Sequence-level amalgamation}. For the characteristic of transformers, we simply extend the student sequences to reach the same length as the concatenated teacher ones. In this way, the student can be supervised by all the teachers explicitly, instead of learning a fixed-size hint with noise and loss of information, which has been widely adopted in previous KA methods. This approach turns out to be crucial, especially when the student has not been initialized by special pre-training. Additionally, as the length of the concatenated sequences increases linearly with the number of tasks, we further propose a compression algorithm to decrease the length to constant while preserve informative vision tokens.

\textit{Task-level amalgamation}.
Similar to the set prediction procedure that uses ground truth labels to train DETR models, we introduce task-level amalgamation to transfer the dark knowledge through soft targets. Every object predicted by the student is matched to a unique responsible teacher object during the training process. Then the task-level amalgamation loss is computed efficiently for all the matched teacher-student pairs weighted by the teacher confidence.

The main contribution of this work is a novel knowledge amalgamation scheme for transformer-based object detection models via sequence-level and task-level amalgamation. Extensive experiments have been conducted on Pascal VOC~\cite{everingham2010pascal} and COCO~\cite{lin2014microsoft} datasets, which have revealed encouraging results: 
the sequence-level amalgamation significantly boosts the performance of transformer-based students, while the previous feature aggregation methods have a negative impact;
moreover, the students turn out to be competent in learning amalgamated knowledge, as they have achieved superior or at least comparable performance to those of the teachers even without spending enormous time for pre-training.

\section{Related Work}
In this section, we briefly review prior work related to our method in the fields of vision transformers and model reusing.

\subsection{Vision Transformers}
The transformer~\cite{NIPS2017_attention}, an encoder-decoder architecture, relies entirely on global attention mechanism and multilayer perceptron for machine translation and have achieved the state of the art performance. With the help of the global attention mechanism, relation between long range tokens can be easily uncovered. The following transformer based architectures~\cite{devlin2018bert,radford2018improving,radford2019language,brown2020gpt} are pre-trained firstly on large-scale corpus and then fine-tuned on a wide range of NLP tasks. Due to the high performance and flexibility, transformers are gradually becoming universal models in NLP.

Inspired by the great success in NLP, transformer-based methods are emerging rapidly in many CV tasks. In image classification, ViT~\cite{dosovitskiy2021vit} directly inherits from BERT~\cite{devlin2018bert} by splitting an image into patches as its input sequence. It presents excellent results compared with CNNs~\cite{szegedy2015going}. Based on ViT,~\cite{touvron2020deit,wu2021cvt,zhou2021deepvit,yuan2021tokenvit} surpass the CNN models with even fewer parameters.

In the field of object detection, different from CNN based methods~\cite{girshick2014rich,girshick2015fast,ren2015faster,liu2016ssd,lin2017focal,cai2018cascade,bochkovskiy2020yolov4} that relay on hand-crafted procedures such as predefining anchors and non-maximal suppression, DETR~\cite{carion2020detr} uses a transformer to predict objects in parallel, and is trained end-to-end by set prediction loss inspired from~\cite{stewart2016end}. Although the performance of DETR is compatible with the common CNNs (\eg, Faster RCNN\cite{ren2015faster}), it suffers from extreme training time. UP-DETR~\cite{dai2020up-detr} proposes an unsupervised pre-training method for DETR which significantly boosts the performance of DETR, and outperforms CNN based models with fewer parameters. A more recent work~\cite{zhu2020deformable} proposes a deformable attention mechanism to fasten the convergence speed of DETR.

Despite the great potential of transformers,~\cite{devlin2018bert,radford2018improving,radford2019language,brown2020gpt,dosovitskiy2021vit,dai2020up-detr} have noticed that transformer based models have a common character that their performances heavily depend on proper initialization points. Nonetheless, it is not necessary for the student in our settings.

\subsection{Model Reusing}
As a growing number of well-trained deep models are available in the Internet, reusing them for model compression and knowledge transfer has become increasingly prevalent in recent years. Hinton \etal~\cite{hinton2015distilling} first propose knowledge distillation that the knowledge of a cumbersome teacher can be distilled to a lightweight student through soft targets predicted by the teacher. They point out that the soft targets which reveal how teachers tend to generalize, regularize the student. To sufficiently distill knowledge, FitNets~\cite{romero2014fitnets} additionally use intermediate features (also known as hints) of the teacher and the following works~\cite{zagoruyko2016paying,tian2019contrastive,song2021tree} extract deeper level of information from the intermediate layers of teachers to supervise the student in different aspects.

However, these methods are only for reusing image classification models~\cite{szegedy2015going,zagoruyko2016wide,howard2017mobilenets,tan2019efficientnet}. As a more challenging CV task, there are a variety of architectures designed for object detection. For example, there are one-stage~\cite{liu2016ssd,lin2017focal} and two-stage~\cite{ren2015faster,cai2018cascade} approaches; and also anchor-based~\cite{liu2016ssd,ren2015faster} and anchor-free~\cite{yang2019reppoints,tian2019fcos} approaches. Initially, Chen \etal~\cite{chen2017learning} and Li \etal~\cite{li2017mimicking} propose distillation methods for faster R-CNN~\cite{ren2015faster}. In~\cite{zhu2019mask}, Zhu \etal~use mask guided feature to solve foreground-background imbalance problem when distilling a single shot detector~\cite{liu2016ssd}. However, Zhang \etal~\cite{zhang2020improve} argue that most KD methods designed for image classification have failed on object detection tasks because of the imbalance between foreground and background pixels, and lack of relation between objects. With attention-guided and non-local distillation, their method is applied to various object detection models, including one-stage and two-stage architectures, and has achieved general improvement. Recently, Dai \etal~\cite{dai2021general} utilize valuable relationship between general instances instead of heavily relying on the ground truth for plenty of object detection models. Their results show that slim students can perform even better than teachers.

In NLP, researchers~\cite{sanh2019distilbert,sun2019patient,jiao2019tinybert} have explored methods to distill a pre-trained transformer-based language model to a tiny one. Particularly, TinyBERT~\cite{jiao2019tinybert} jointly uses intermediate attention masks and hidden states to supervise the student. They discover that the intermediate supervision somehow initializes the student which plays an important role in training a transformer based model. 

Furthermore, researchers~\cite{shen2019customizing,shen2019amalgamating,ye2019amalgamating,ye2019student,li2020knowledge} are starting to investigate a novel model reusing task, named knowledge amalgamation. As an extension of KD, the knowledge of more than one teacher is transferred to a compact and multi-talented student in the KA setting. Specifically,~\cite{shen2019customizing,shen2019amalgamating,ye2019amalgamating} focus on the situation where teachers are trained on homogeneous tasks, \eg, multiple image classification tasks, while~\cite{ye2019student,li2020knowledge} are proposed for teachers trained on disparate tasks like semantic segmentation and depth estimation.
In particular, to amalgamate intermediate teacher features,~\cite{shen2019amalgamating} develops an encoder-decoder structure. Luo \etal~\cite{luo2019knowledge} adopt common feature learning to project features of all the teachers and the student close to each other. These CNN-based KA approaches share a common strategy that the student requires a fixed-sized hint (generated mostly by projection), which suffers from extra learning burden and loss of information. Nowadays, most KA works are proposed in the realm of CNNs, apart from~\cite{jing2021amalgamating} for graph neural networks. 

To the best of our knowledge, none of existing KA methods have explored a solution for reusing object detectors nor vision transformers.

\section{Method}
\begin{figure*}[!t]%
\centering%
\includegraphics[width=\textwidth]{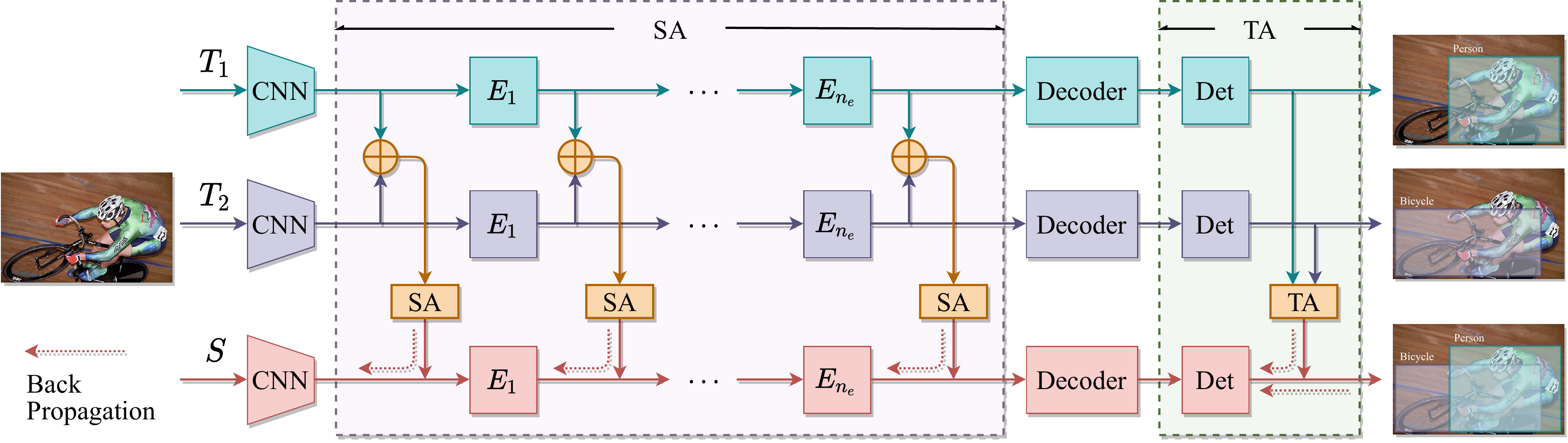}%
\caption{%
The overall workflow of our proposed knowledge amalgamation method for transformer-based object detectors is shown in the two-teacher case. Our method aims to transfer knowledge from several well-trained teachers to one compact yet multi-talented student, which is explicitly capable of detecting all the categories taught by given teachers. As shown in the figure, our method is composed of two separate components. (1) Sequence-level amalgamation (SA): for the CNN backbone and all encoder layers, the intermediate sequence of the student is supervised by concatenated (and compressed) outputs of all corresponding teacher sequences. (2) Task-level amalgamation (TA): the student learns heterogenous detection tasks by mimicking the soft targets predicted by teachers. The gradient flows are shown as the dotted lines from our KA modules and the ground truth supervision.%
}
\label{fig:workflow}%
\end{figure*}

The overall workflow of our proposed method is shown in Figure~\ref{fig:workflow} in the two-teacher case, in which knowledge of well-trained teachers is transferred to the student by two separated components: sequence-level amalgamation~(SA) and task-level amalgamation~(TA). In our setting, all the teachers and the student are transformer-based object detection models, specifically DETR. We first briefly introduce DETR and KA, then delineate the two parts of our proposed KA method respectively in Section~\ref{sec:1} and \ref{sec:2}.

\subsection{Preliminaries}
\subsubsection{Object Detection with Transformers}
DETR flattens a two-dimensional output feature map of a CNN backbone to the sequence with $n$ tokens $X=(x_1, \ldots, x_n) \in \mathbb{R}^{n\times d}$, where each token $x_i$ is a $d$-dimensional embedding vector. Afterward, the sequence is fed into an encoder-decoder structure, \ie, the transformer. Finally, a task header takes the decoded sequence to predict a set of $m$ objects $\hat{Y} = \{\hat{y}_i\}_{i=1}^m$, each of which is a binding of two terms $\hat{y} = (\hat{b}_i, \hat{p}_i)$: the bounding box position $\hat{b}_i \in \mathbb{R}^4$; and the bounding box category distribution $\hat{p}_i \in \mathbb{R}^{|C|}$, where $C$ is the set of all the categories.

The transformer of DETR is composed of only two modules: the multi-head attention mechanism~(MHA) and the multilayer perceptron~(MLP). We will briefly review these two building blocks for the convenience of further discussion.

The MHA module can be treated as the linear projection of concatenated outputs of several dot-product attention modules. Equivalent to the original formula~\cite{NIPS2017_attention}, we rewrite the output of MHA in the matrix form\footnotemark
\begin{equation}
    \text{MHA}(Q, K, V) = \sum_{i=1}^{h} \sigma\left( \frac{A_i}{\sqrt{d_k}}\right)V W_i^{VO} \text{,}
    \label{eq:mha}
\end{equation}
where $Q\in\mathbb{R}^{n_q\times d_{\text{model}}}$, $K\in\mathbb{R}^{n\times d_{\text{model}}}$ and $V\in\mathbb{R}^{n\times d_{\text{model}}}$ are queries, keys and values; $h$ is the number of heads; $A_i$ is the attention matrix computed from $Q$, $K$ and projection matrices; $W_i^{VO}$ is a combination of projection matrices; $d_k = d_{\text{model}}/h$; and $\sigma(\cdot)$ represents softmax operation applied to each row of the input matrix. As a special case, the MHSA module has the same inputs $Q=K=V$. Although the MHA module creates information flow between different tokens, it can still be natively extended to spliced sequences as the shape of all projection matrices is independent of the sequence's length, which is vital for constructing our SA method.

\footnotetext{See Appendix~\ref{appendix:1} for detailed derivation.}

The MLP layer maps a single token $x\in\mathbb{R}^d$ to $y = \text{MLP}(x)$ with the same dimension as $x$, and as a result, $y$ is independent of all other tokens.

\subsubsection{Knowledge Amalgamation}
We define knowledge amalgamation mathematically in the field of object detection as follows. To begin with, let $\mathcal{D}=\{I_i\}_{i=1}^L$ be a dataset of $L$ images. For image $I_i$, we have a set of annotations $Y_i=\{y_i^j\}_{j=1}^{a_i}$, where $a_i$ is the number of annotations for $I_i$. The dataset has been annotated with $|C|$ categories in total.
Here an annotation $y=(b, c)$ is a binding of bounding box position $b\in \mathbb{R}^4$, and its category $c\in C$ encoded as an integer, $C=\{1, \ldots, |C|\}$.

With these notations, we define a \textit{task} $\mathcal{T}_t(\cdot;C^t)$ as a mapping between two image annotation sets parameterized by a partition $C^t \subseteq C$. Further, the annotation set of image $I_i$ is constructed under the task $\mathcal{T}_t$ as $Y_i^t = \mathcal{T}_t(Y_i;C^t) \subseteq Y_i$ such that all annotations with category $c\in C^t$ are preserved and those with $c\notin C^t$ are discarded. In other words, a task $\mathcal{T}_t$ built upon an object detection dataset will filter all the annotations with unwanted categories.

In our KA setting, $N$ tasks $\mathcal{T}=\{\mathcal{T}_t(\cdot;C^t)\}_{t=1}^N$ are built for the same detection dataset, and $N$ teachers $\{T_t\}_{t=1}^N$ have been well-trained on each task accordingly. Without loss of generality, we assume that for any two tasks $\mathcal{T}_i$, $\mathcal{T}_j\in\mathcal{\mathcal{T}}$, the intersection of their partitions $C^i \cap C^j = \varnothing$, and the student is trained with all categories $C = \bigcup_{i=1}^N C^i$.

\subsection{Sequence-level Amalgamation}
\label{sec:1}
\begin{figure}[!t]%
\centering%
\includegraphics[width=\linewidth]{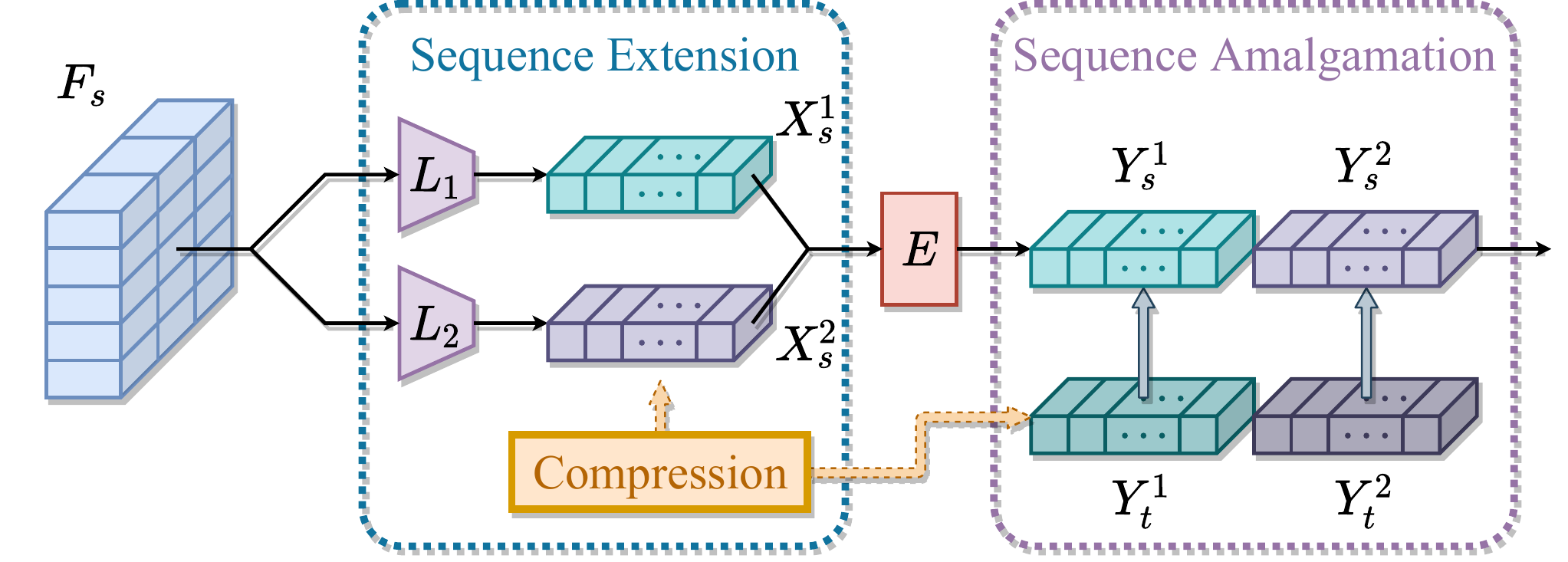}%
\caption{%
    Illustration of sequence extension and sequence-level amalgamation for the student model. We duplicate the linear projection layer (after the CNN backbone) to generate the extended student sequences (\eg, $X_S = X_s^1 \oplus X_2^2$). In this way, the student sequences (\eg, $Y_S = Y_s^1 \oplus Y_s^2$) can be directly supervised by the corresponding teacher sequences (\eg, $Y_T = Y_t^1 \oplus Y_t^2$). Additionally, the sequence compression module is utilized to decrease the computation cost, as explained in Section~\ref{sec:seq-compression}.
}%
\label{fig:seq-exten}%
\end{figure}

This section proposes the \textit{sequence-level amalgamation}, where the student learns amalgamated knowledge through intermediate sequence supervision. To simplify the derivation process, we take the case of two teachers as an example.

\subsubsection{Sequence-level Amalgamation Loss}
\label{sec:SA-loss}
Given two teachers $T_1$ and $T_2$, and with the assumption that all the sequences share the same embedding dimension $d$, we can rewrite $X_t^i\in\mathbb{R}^{n\times d}$ fed into layer $L$ of all the teachers in matrix form as
\begin{equation*}
    X_T = X_t^1 \oplus X_t^2 =
    \begin{bmatrix}
        X_t^1 \\
        X_t^2 \\
    \end{bmatrix}
    \text{,}
\end{equation*}
where the operation $\oplus$ denotes the sequence concatenation. Therefore, the output sequence can also be written as
\begin{equation*}
    Y_T = Y_t^1 \oplus Y_t^2 =
    \begin{bmatrix}
        L(X_t^1; W_t^1) \\
        L(X_t^2; W_t^2) \\
    \end{bmatrix}
    \text{,}
    \label{eq:teacher-out}
\end{equation*}
where $W_t^1$ and $W_t^2$ are the parameters corresponding to the teacher layer $L$.

However, as the student input sequence $X_s \in \mathbb{R}^{n\times d}$ is half as long as $X_T \in \mathbb{R}^{2n \times d}$, we duplicate the linear projection layer (before the transformer) of the student as shown in Figure~\ref{fig:seq-exten} so that the input sequence to the transformer is the concatenation of the two sequences $X_S = X_s^1 \oplus X_s^2$, which is as long as $X_T$. After that, $X_S$ is fed into the encoder of transformers. The output $Y_S$ from student layer $L$ can be directly supervised by $Y_T$ (\eg, the Euclidean distance) without any extra projection module (used by previous KA methods). Note that for DETR architecture, the extension of the encoder sequence will not influence the decoder part.

In this way, the \textit{sequence-level amalgamation loss} for all layers (the linear projection layer after the CNN backbone can also be included) in the encoder boundary is defined as
\begin{equation}
    \mathcal{L}_\text{seq} = \sum_{l=1}^{n_e} \mathcal{L}_\text{seq}^l = \frac{1}{N} \sum_{l=1}^{n_e} \|Y_S^l - T_T^l\|_F^2 \text{,}
    \label{eq:SA}
\end{equation}
where $n_e$ is the number of encoder layers; $N$ is the number of teachers.

In order to avoid interference within $X_s^i$ during the forward propagation, \ie, the computation of $Y_s^i$ depends on other input sequences $X_s^j$ ($i \neq j$), we use an attention mask in MHSA layers to cut off the information flow within these sequence parts.

Firstly, given concatenated input sequence $X_S$, based on Equation~\ref{eq:mha}, the output sequence $Y_S$ (we omitted the summation of all the heads for convenience) from the MHSA layer is
\begin{equation}
    Y_S = \sigma\left(
        \begin{bmatrix}
            A_{11}^s & A_{12}^s \\
            A_{21}^s & A_{22}^s \\
        \end{bmatrix} / \sqrt{d_k}
    \right)
    \begin{bmatrix}
        X_s^1 W_s^{VO} \\
        X_s^2 W_s^{VO} \\
    \end{bmatrix} \text{.}
    \label{eq:concated-output}
\end{equation}

Here, the sub-matrices $A_{12}^s$ and $A_{21}^s$ (computed from both $X_s^1$ and $X_s^2$) cause interference. Thus, by masking out the off-diagonal sub-matrices in Equation~\ref{eq:concated-output} by $-\infty$, the decoupled form is eventually achieved as
\begin{equation}
    Y_S =
    \begin{bmatrix}
        \sigma(A_{11}^s / \sqrt{d}) X_s^1 W_s^{VO} \\
        \sigma(A_{22}^s / \sqrt{d}) X_s^2 W_s^{VO} \\
    \end{bmatrix} \text{.}
    \label{eq:seq-decoupled-form}
\end{equation}

It is worth pointing out that, on the one hand, by masking out cross attention terms, the output sequence of every encoder layer is the concatenation of decoupled form since MLP layers process each token independently. On the other hand, with the help of the attention mask, the computational complexity of the MHA layer is, therefore, decreased from $\mathcal{O}(N^2)$ to $\mathcal{O}(N)$ with respect to the number of teachers.

\subsubsection{Gradient Analysis for Sequence-level Supervision}
\label{sec:seq_amg_analysis}
As shown in Equation~\ref{eq:seq-decoupled-form}, we have achieved the decoupled form, where each part of the student sequence is supervised individually by the corresponding teacher in the forward propagation. In this section, we compare our proposed SA approach and previous feature (or sequence) aggregation approach from the perspective of gradients to figure out how the student parameters are updated during the backward propagation. For the sake of simplicity, we focus on student layer $L$ with parameter $W_s$.

Here, we define the sequence aggregation (SAG) approach (usually adopted in previous KA methods) with direct supervision between original student sequence $Y_s \in\mathbb{R}^{n\times d}$ and aggregated teacher sequence $Y_a \in\mathbb{R}^{n\times d}$ as
\begin{equation}
    \mathcal{L}_\text{SAG} = \|Y_s - Y_a\|_F^2 \text{,}
    \label{eq:SAG}
\end{equation}
where $Y_a = \frac{1}{N} \sum_{i=1}^N Y_t^i W_a^i$ and $W_a^i\in\mathbb{R}^{d \times d}$ is the common space projection matrix.

By taking the derivative of the SA loss in Equation~\ref{eq:SA} with respect to $W_s$, we have
\begin{equation}
    \frac{\partial \mathcal{L}_\text{SA}^l}{\partial W_s} = \frac{2}{N} \sum_{i=1}^{N} (Y_s^i - Y_t^i) \frac{\partial L(X^i; W_s)}{\partial W_s} \text{,}
    \label{eq:SA-derivative}
\end{equation}
where the output teacher sequence $Y_t^i = L(X^i; W_t^i)$ and the according student sequence $Y_s^i = L(X^i; W_s)$ depend on the same input $X^i$.

Similarly, the derivative of the SAG loss in Equation~\ref{eq:SAG} with respect to $W_s$ is
\begin{equation}
    \frac{\partial \mathcal{L}_\text{SAG}^l}{\partial W_s} = \frac{2}{N} \sum_{i=1}^N (Y_s - Y_t^i W_a^i) \frac{\partial L(X; W_s)}{\partial W_s} \text{.}
    \label{eq:SAG-derivative}
\end{equation}

In both Equation~\ref{eq:SA-derivative} and~\ref{eq:SAG-derivative}, the student is guided by several teachers simultaneously. However, since $W_a^i$ is learned from scratch, which is different from the optimal projection matrix $\tilde{W}_a^i$, it brings an error gradient
\begin{equation*}
    G_\text{err} = \frac{2}{N} \sum_{i=1}^N Y_t^i \varDelta_a^i \frac{\partial L(X; W_s)}{\partial W_s} \text{,}
\end{equation*}
where $\varDelta_a^i = \tilde{W}_a^i - W_a^i$. In consequence, $\varDelta_a^i$ leads to the error direction. Because of this, throughout the whole training procedure, the SAG approach suffers from noisy gradients. In Section~\ref{exp:ablation-agg}, we compare the performance of both sequence-level amalgamation approaches, which reveals that SAG even makes the student worse.

Above all, utilizing our proposed SA method, the intermediate layers of the student are trained from compatible supervision since each part of the student sequence is calculated independently during the forward propagation. Furthermore, the student learns the common knowledge without disturbance during the backward propagation.

\subsubsection{Sequence Compression}
\label{sec:seq-compression}
\begin{figure*}[!t]%
\centering%
\subfloat[Isometric down-sampling]{%
    \includegraphics[width=0.32\linewidth]{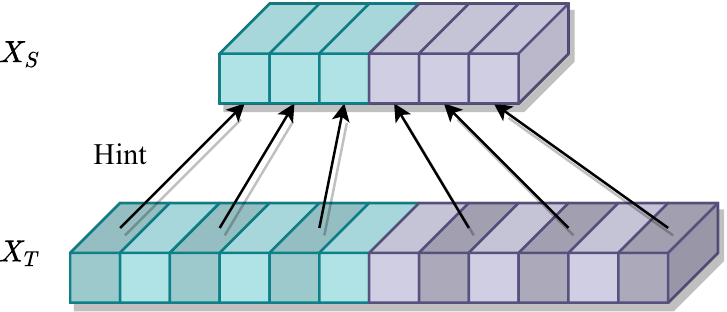}%
    \label{fig:comp-a}%
}%
\hfil%
\subfloat[Random down-sampling]{%
    \includegraphics[width=0.32\linewidth]{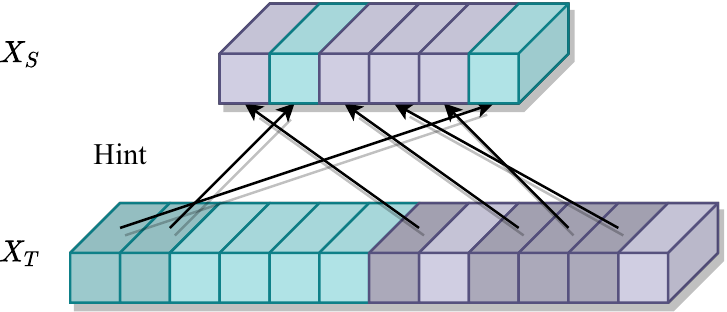}%
    \label{fig:comp-b}%
}%
\hfil%
\subfloat[Redundancy-based compression]{%
    \includegraphics[width=0.32\linewidth]{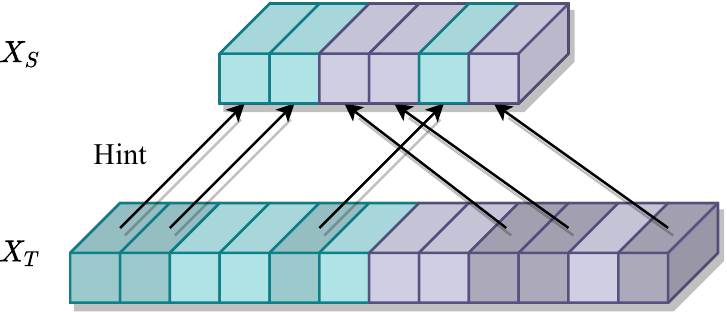}%
    \label{fig:comp-c}%
}%
\caption{Illustration of different sequence compression methods. The student sequences are supervised by partial teacher hints, compressed based on various sequence compression strategies, \ie, index set $P_\text{slim}$. However, we have kept the integrity of vision tokens (they are either preserved or discarded) to avoid introducing noise as the SAG approach.}%
\label{fig:comp}%
\end{figure*}

\begin{algorithm}[!t]
\caption{Compressing extended student sequence}
\label{alg:1}
\begin{algorithmic}[1]
    \Require $X_T$: concatenated teacher sequence; $X_S$: extended student sequence; $N$: number of teachers; $n$: length of original student sequence.
    \Ensure $P_\text{slim}$: set of reserved token indices with $|P_\text{slim}| = n$
    \State $P_\text{slim} \gets \varnothing$
    \For{$i \gets 1:n$}
        \State $t_\text{keep} \gets 0$; $r_\text{min} \gets \infty$
        \For{$t \gets 1:N$}
            \State $r \gets R(x_t^i|X_T)$ \label{alg:line1}
            \If{$r < r_\text{min}$}
                \State $t_\text{keep} \gets t$; $r_\text{min} \gets r$
            \EndIf
        \EndFor
        \State \# Keep the $i^\text{th}$ token of teacher $t_\text{keep}$
        \State $i_\text{keep} \gets (t_\text{keep} - 1)n + i$
        \State $P_\text{slim} \gets \{i_\text{keep}\} \cup P_\text{slim}$
    \EndFor
\end{algorithmic}
\end{algorithm}

In Equation~\ref{eq:seq-decoupled-form}, we decrease the computation cost to $\mathcal{O}(N)$ regarding the number of tasks, which is linear growth. As shown in Section~\ref{exp:redundancy}, the extended student sequence has plenty of redundant tokens. Therefore, the computation cost can be further decreased without significant performance degradation.

Given a sequence $X=(x_1, \ldots, x_n)$ and an index set $P_n = \{1, \ldots, n\}$, the compression process is to find a permutation index set $P \subset P_n$ with cardinality $m < n$, so that $[X]_P = (x_{p_1}, \ldots, x_{p_m})$ is the compressed sequence selected by $P$. As Appendix~\ref{appendix:3} has demonstrated that transformers have the invariance property under the permutation of input tokens, we assume $p_1 < \cdots < p_m$ for convenience.

To compress such sequences, we propose some kinds of sequence compression strategies in Figure~\ref{fig:comp}. One most straightforward strategy is using isometric down-sampling, as shown in Figure~\ref{fig:comp-a}, where tokens are sampled alternately from every teacher. However, this strategy treats all tokens equally and fails to filter redundant ones as the sequence information of each teacher varies for different inputs.

In order to compress the sequences according to their redundancy, we first define the redundancy of tokens mathematically and then propose a redundancy-based sequence compression algorithm, illustrated in Figure~\ref{fig:comp-c}.

\begin{definition}[Token redundancy]
    The redundancy $R(x_i|X)$ of a token $x_i$ in the sequence $X$ is the mean of the $i^{\text{th}}$ row of the sequence similarity matrix $S\in\mathbb{R}^{n\times n}$, computed as
    \begin{equation}
        R(x_i|X) = \frac{1}{n} \sum_{j=1}^{n}S_{i,j} \text{.}
        \label{eq:redundancy}
    \end{equation}
\end{definition}

Here, the similarity matrix $S$ is calculated by $S = \hat{X} \hat{X}^\top$, where $\hat{X}$ is the normalized sequence $\hat{X} = (\hat{x}_1, \ldots, \hat{x}_n)$ and $\hat{x}_i = x_i/\|x_i\|_2$. 

The redundancy-based sequence compression strategy is shown in Algorithm~\ref{alg:1}, which generates a set of reserved token indices $P_\text{slim}$. Selected by $P_\text{slim}$, the length of the extended student sequence is reduced from $Nn$ to $n$ by removing high redundancy tokens while keeping the integrity of objects. To be specific, the algorithm finds a token $x_t^i$ within $\{x_1^i, \ldots, x_N^i\}$ with minimum redundancy in turn and eventually uses these low redundancy tokens to reconstruct an informative sequence. As such, the extended student sequence is compressed to $\mathcal{O}(1)$ with regard to the task number.

The sequence compression procedure is conducted right after the linear projection module demonstrated in Figure~\ref{fig:seq-exten}, and the compressed sequence is fed into the following student transformers. Besides, the output teacher sequence $Y_T^l$ from encoder layer $l$ is identically compressed to $[T_T^l]_{P_\text{slim}}$ according to $P_\text{slim}$ to reach the same length as the student sequence $Y_S^l$.

At the training stage, we use the concatenated teacher sequences $X_T$ to guide the compression procedure instead of $X_S$ because the student has not converged yet. Furthermore at the evaluating stage, the student sequence is compressed based on $X_S$ only.

\subsection{Task-level Amalgamation}
\label{sec:2}
Collaborated with SA, in this section, we introduce the \textit{task-level amalgamation} method where the student learns to detect all sorts of objects by mimicking soft targets predicted by several teachers.

\subsubsection{Amalgamating the Predictions}
Suppose that a teacher predicts $m$ targets $Y_t = \{y_t^i\}_{i=1}^m$, where each target $y_t^i = (b_t^i, p_t^i)$ is a binding of a bounding box $b_t^i \in \mathbb{R}^4$ and its probability distribution $p_t^i \in \mathbb{R}^{|C^t|}$. As all the teachers are trained on disjoint category subsets $\{C^1, \ldots, C^N\}$, we pad the bounding box prediction $p$ to $\tilde{p} \in \mathbb{R}^{|C|}$ to amalgamate the predicted objects compatibly. For each category $c \in \bigcup_{n=1}^N C^n$, the corresponding component of $\tilde{p}$ is
\begin{equation*}
    \tilde{p}_c = \begin{cases}
        p_c & \mbox{if } c \in C^t \text{,}\\
        0 & \mbox{otherwise} \text{.}
    \end{cases}
\end{equation*}
During the following discussion, we replace the padded teacher prediction $\tilde{p}$ with $p$ to shorthand the notation. With the padded probability distribution, all the teacher predictions can be amalgamated as $Y = \bigcup_{t=1}^N Y_t$ compatibly.

Meanwhile, the student predicts $m$ targets $\hat{Y} = \{\hat{y}_i\}_{i=1}^m$, where each target is also a binding of $\hat{b}_i$ and $\hat{p}_i \in \mathbb{R}^{|C|}$.

\subsubsection{Target Set Matching}
In general, the target sets predicted by all the teachers and the student have the same cardinality, a hyperparameter to DETR. Thus, teachers predict more targets than students, which allows every student target $\hat{y}_i$ to be matched with $y_t^j$ for sufficient supervision.

Let $\sigma(\cdot)$ denotes a teacher-student match function (following the notation in~\cite{carion2020detr}) that student target $i$ is matched to teacher target $\sigma(i)$, and $\mathfrak{S}$ is the set of all the match functions.

To find a optimal teacher-student target match $\hat{\sigma}\in \mathfrak{S}$, we minimize the cost function
\begin{equation}
    \mathcal{L}_\sigma = \sum_{i}^{m} \mathcal{L}_\text{match}(y_{\sigma(i), \hat{y}_i}) \text{,}
\end{equation}
where the pairwise matching cost $\mathcal{L}_\text{match}$ is analogous to~\cite{carion2020detr}
\begin{equation}
\begin{split}
    \mathcal{L}_\text{match}(y_{\sigma(i)},\!\hat{y}_i)\!= &\alpha_1 D_{KL}(p_{\sigma(i)} \| \hat{p}_i)
    \!+\!\alpha_2 \mathcal{L}_\text{box}(b_{\sigma(i)},\!\hat{b}_i) \\
    &\!-\!\alpha_3 \mathcal{L}_\text{conf}(p_{\sigma(i)}) \text{.}
\end{split}
\label{eq:loss-match}
\end{equation}
Here, $D_{KL}(P \| Q)$ is the Kullback-Leibler divergence from distribution $Q$ to $P$; $\mathcal{L}_\text{box}$ is weighted sum of $\ell_1$ loss and generalized IoU loss following~\cite{carion2020detr}; $\mathcal{L}_\text{conf}(p)$ is the maximum component of distribution $p$ apart from background terms, which makes priority for matching teacher targets with high confidence; and finally, $\alpha_1$, $\alpha_2$ and $\alpha_3$ are the weighting parameters.

\subsubsection{Task-level Amalgamation Loss}
After finding the optimal matching function $\hat{\sigma}$, all the student targets can be supervised by the matched teacher targets. The \textit{Hungarian} distillation loss, \ie, task-level amalgamation loss, is defined as
\begin{equation}
    \begin{split}
    \mathcal{L}_\text{task} = & \sum_{i=1}^n \mathcal{L}_\text{conf}(p_{\hat{\sigma}(i)}) [\beta_1 D_{KL}(p_{\hat{\sigma}(i)} \| \hat{p}_i)\\
    & + \beta_2 \mathcal{L}_\text{box}(b_{\hat{\sigma}(i)}, \hat{b}_i) ] \text{,}
\end{split}
\label{eq:loss-task}
\end{equation}
where $\beta_1$ and $\beta_2$ are the weighting parameters. Notably, we use the confidence $\mathcal{L}_\text{conf}(p_{\hat{\sigma}(i)})$ of a teacher prediction $\hat{\sigma}(i)$ for balancing the foreground and background loss as the pairwise supervision weight.

In practice, due to the Hungarian algorithm has the computation complexity of $\mathcal{O}(n^3)$, we filter out teacher targets with low confidence to speed up the training process.

\subsection{Final Amalgamation Loss}
The final amalgamation loss is similar to most KD work which is composed of three terms: (1) KL divergence of teacher soft target and student prediction, (2) intermediate level supervision, and (3) directly training loss with labeled data. Although in image classification, the third term can be omitted as the soft targets of teachers are strong enough to train the student, it is crucial when training a detection model, as shown in Section~\ref{exp:GT}.

In conclusion, the final amalgamation loss is the weighted sum of SA, TA and directly training loss as
\begin{equation}
    \mathcal{L} = \lambda_\text{seq} \mathcal{L}_\text{seq} + \lambda_\text{task} \mathcal{L}_\text{task} + \lambda_d \mathcal{L}_d \text{,}
\label{eq:loss-final}
\end{equation}
where $\lambda_\text{seq}$, $\lambda_\text{task}$ and $\lambda_d$ are the weighting parameters and $\mathcal{L}_d$ is directly training loss same as~\cite{carion2020detr}.

\section{Experiments}
\begin{table*}[!t]
\centering
\setlength{\tabcolsep}{4pt}
\caption{Knowledge Amalgamation Results on Object Detection Datasets}
\resizebox{\textwidth}{!}{\begin{tabular}{l *{5}{r} *{5}{r}}
    \toprule
    \multicolumn{1}{c}{\textbf{Dataset}}
    & \multicolumn{5}{c}{\textbf{PASCAL VOC}}
    & \multicolumn{5}{c}{\textbf{MS COCO}} \\
    \cmidrule(lr){1-1}\cmidrule(lr){2-6} \cmidrule(lr){7-11}
    \multicolumn{1}{c}{\textbf{Model}}
    & \multicolumn{1}{c}{AP}
    & \multicolumn{1}{c}{AP$_{50}$}
    & \multicolumn{1}{c}{AP$_{75}$}
    & \multicolumn{1}{c}{\#param.}
    & \multicolumn{1}{c}{FLOPs}
    & \multicolumn{1}{c}{AP}
    & \multicolumn{1}{c}{AP$_{50}$}
    & \multicolumn{1}{c}{AP$_{75}$}
    & \multicolumn{1}{c}{\#param.}
    & \multicolumn{1}{c}{FLOPs} \\
    \midrule
    Teacher$_1$
    & 46.15 & 74.19 & 47.80 & 41.3M & 12.7G
    & 36.65 & 58.06 & 47.97 & 41.3M & 89.6G \\
    Teacher$_2$
    & 51.63 & 79.40 & 56.43 & 41.3M & 12.7G 
    & 39.75 & 59.35 & 42.15 & 41.3M & 89.6G \\
    Ensemble
    & 48.89 & 76.80 & 52.12 & 82.6M & 25.4G 
    & 38.20 & 58.71 & 40.06 & 82.6M & 179.2G \\
    \midrule
    Raw
    & 44.94 & 73.52 & 47.16 & 41.3M & 12.7G
    & 32.64 & 52.41 & 34.22 & 41.3M & 89.6G \\
    Raw$_\text{ext}$
    & 43.49 & 71.96 & 45.17 & 41.8M & 14.2G
    & 32.51 & 51.86 & 33.91 & 41.8M & 107.15G \\
    Student$_{\text{slim}}$
    & 49.86 (\Blue{+4.92}) & 78.16 (\Blue{+4.64}) & 52.17 (\Blue{+5.01}) & 41.8M & 12.8G
    & 35.91 (\Blue{+3.27}) & 56.30 (\Blue{+3.89}) & 37.70 (\Blue{+3.48}) & 41.8M & 90.1G \\
    Student
    & 49.94 (\Blue{+5.00}) & 78.91 (\Blue{+5.39}) & 52.82 (\Blue{+5.66}) & 41.8M & 14.2G
    & 37.08 (\Blue{+4.44}) & 56.96 (\Blue{+4.55}) & 39.02 (\Blue{+4.80}) & 41.8M & 107.15G \\
    \midrule
    Raw$_{\text{UP}}$
    & 51.02 & 79.63 & 53.36 & 41.3M & 12.7G
    & 35.56 & 55.56 & 36.90 & 41.3M & 89.6G \\
    Student$_{\text{UP}}$
    & 51.64 (\Blue{+0.62}) & 80.06 (\Blue{+0.43}) & 54.76 (\Blue{+1.40}) & 41.8M & 14.2G
    & 37.72 (\Blue{+2.16}) & 58.18 (\Blue{+2.62}) & 39.16 (\Blue{+2.26}) & 41.8M & 107.15G \\
    \bottomrule
\end{tabular}}
\label{table:result}
\end{table*}

\subsection{Settings}
\subsubsection{Dataset}
Two widely-used object detection datasets are adopted to evaluate our proposed method as benchmarks, including PASCAL VOC and COCO. There are 20 categories annotated in VOC and 80 categories in COCO. Like UP-DETR~\cite{dai2020up-detr}, we jointly use the \textit{trainval07+12} partitions (about 16.5k images) as the training dataset and the \textit{test07} partition (about 5k images) as the validation dataset. We use the \textit{train2017} partition (about 118k images) as the training dataset and the \textit{val2017} partition (5k images) as the validation dataset for the COCO dataset.

\begin{table}[!t]
\centering
\caption{Partition Scheme of VOC Categories}
\setlength{\tabcolsep}{1.8pt}
\resizebox{\linewidth}{!}{\begin{tabular}{ll *{10}{l}}
    \toprule
    \multicolumn{1}{c}{\textbf{Settings}} & \multicolumn{1}{c}{\textbf{Tasks}} & \multicolumn{5}{c}{\textbf{Categories}} \\
    \midrule
    \multirow{4}{*}{Two Teachers} 
    & \multicolumn{1}{c}{\multirow{2}{*}{$\mathcal{T}_1$}}
    & Aeroplane & Bicycle & Bird & Boat & Bottle \\
    &  & Bus & Car & Cat & Chair & Cow \\
    \cmidrule(l){2-7}
    & \multicolumn{1}{c}{\multirow{2}{*}{$\mathcal{T}_1$}}
    & Dining Table & Dog & Horse & Motorbike & Person \\
    &  & Potted Plant & Sheep & Sofa & Train & TV Monitor \\
    \midrule
    \multirow{4}{*}{Four teachers}
    & \multicolumn{1}{c}{$\mathcal{T}_1$} & Aeroplane & Bicycle & Bird & Boat & Bottle \\
    & \multicolumn{1}{c}{$\mathcal{T}_2$} & Bus & Car & Cat & Chair & Cow \\
    & \multicolumn{1}{c}{$\mathcal{T}_3$} & Dining Table & Dog & Horse & Motorbike & Person \\
    & \multicolumn{1}{c}{$\mathcal{T}_4$} & Potted Plant & Sheep & Sofa & Train & TV Monitor \\
    \bottomrule
\end{tabular}}
\label{table:voc-split}
\end{table}

\begin{table}[!t]
\centering
\caption{Partition Scheme of COCO Categories}
\setlength{\tabcolsep}{4pt}
\resizebox{\linewidth}{!}{\begin{tabular}{ll *{10}{l}}
    \toprule
    \multicolumn{1}{c}{\textbf{Settings}} & \multicolumn{1}{c}{\textbf{Tasks}} & \multicolumn{10}{c}{\textbf{Categories}\footnotemark} \\
    \midrule
    \multirow{8}{*}{Two Teachers}
    & \multicolumn{1}{c}{\multirow{4}{*}{$\mathcal{T}_1$}}
    &   1 &  2 &  3 &  4 &  5 & 10 & 11 & 16 & 17 & 18 \\
    && 19 & 20 & 27 & 28 & 31 & 34 & 35 & 36 & 37 & 38 \\
    && 44 & 46 & 47 & 52 & 53 & 54 & 55 & 56 & 62 & 63 \\
    && 64 & 72 & 73 & 74 & 78 & 79 & 80 & 84 & 85 & 86 \\
    \cmidrule(l){2-12}
    & \multicolumn{1}{c}{\multirow{4}{*}{$\mathcal{T}_2$}}
    &   6 &  7 &  8 &  9 & 13 & 14 & 15 & 21 & 22 & 23 \\
    && 24 & 25 & 32 & 33 & 39 & 40 & 41 & 42 & 43 & 48 \\
    && 49 & 50 & 51 & 57 & 58 & 59 & 60 & 61 & 65 & 67 \\
    && 70 & 75 & 76 & 77 & 81 & 82 & 87 & 88 & 89 & 90 \\
    \bottomrule
\end{tabular}}
\label{table:coco-split}
\end{table}

\footnotetext{We only list the category ids of COCO dataset. For detailed information please visit: https://tech.amikelive.com/node-718/what-object-categories-labels-are-in-coco-dataset/}

To construct heterogeneous tasks on the given datasets, we split all the categories into non-overlapping parts of equal size to train the teachers. Specifically, we separate the COCO dataset with approximate super-category distribution so that the difficulty of each task is almost equal. The detailed separation schemes are given in Table~\ref{table:voc-split} and~\ref{table:coco-split} separately for VOC and COCO datasets.

\subsubsection{Implementation}
In our implementation, the DETR object detection model has the same architecture as~\cite{dai2020up-detr}. Besides, all the teachers are first pre-trained as~\cite{dai2020up-detr} and then fine-tuned to heterogeneous detection tasks. The transformer part of a student model is randomly initialized while the CNN backbone (ResNet50) is pre-trained with SwAV~\cite{caron2020unsupervised}.

We implement our method with PyTorch~\cite{paszke2019pytorch} and adopt automatic mixed precision (AMP) training for acceleration and saving memory while maintaining the model performance\footnotemark. AdamW~\cite{loshchilov2017decoupled} is used to optimize the student and KA modules, with the learning rate of $10^{-4}$, weight decay of $10^{-4}$. We use a mini-batch size of 32 on four Nvidia Quadro P6000 (24G) GPUs for training the student.

Due to the computational constraints, the teachers and students on VOC are trained with 200 epochs with relatively smaller image sizes comparing with~\cite{carion2020detr} (the maximum image size in our method is 512 whereas in~\cite{carion2020detr} is 1333). However, as small images will significantly decline the performance on the COCO dataset, we train the teachers on COCO with a maximum image size of 1000 for 150 epochs (for about eight days), while the students are trained with 60 epochs. Nevertheless, experiments have shown competitive results against~\cite{dai2020up-detr} with the help of our KA method.

There are several hyper-parameters involved in our method, including $\alpha_1$, $\alpha_2$ and $\alpha_3$ in Equation~\ref{eq:loss-match} for teacher-student target match; $\beta_1$ and $\beta_2$ in Equation~\ref{eq:loss-task} for the TA loss; $\lambda_\text{seq}$, $\lambda_\text{task}$ and $\lambda_\text{d}$ in Equation~\ref{eq:loss-final} for the final KA loss. For $\alpha_i$ and $\beta_j$, we adopt the values in UP-DETR~\cite{dai2020up-detr}. Besides, we set $\lambda_\text{seq} = \lambda_\text{task} = 1$ and $\lambda_\text{d} = 0.1$.

Finally, to stabilize the training procedure when conducting the sequence-level amalgamation, all the sequences are normalized independently for each channel (the embedding dimension of vision tokens) by channel mean and variance in a mini-batch.

\footnotetext{The code will be available soon.}

\subsubsection{Evaluation}
We evaluate our method with COCO style metrics, including AP (average precision), $\text{AP}_{50}$ (default metric for VOC) and $\text{AP}_{75}$ on the validation datasets.

\subsection{Results and Comparison}
\begin{table}[!t]
\centering
\setlength{\tabcolsep}{2.5pt}
\caption{Knowledge Amalgamation Results in Four Teachers Case}
\resizebox{\linewidth}{!}{\begin{tabular}{l rrrrr}
    \toprule
    \multicolumn{1}{c}{\textbf{Dataset}}
    & \multicolumn{5}{c}{\textbf{PASCAL VOC}} \\
    \cmidrule(lr){1-1}\cmidrule(lr){2-6}
    \multicolumn{1}{c}{\textbf{Model}}
    & \multicolumn{1}{c}{AP}
    & \multicolumn{1}{c}{AP$_{50}$}
    & \multicolumn{1}{c}{AP$_{75}$}
    & \multicolumn{1}{c}{\#param.}
    & \multicolumn{1}{c}{FLOPs} \\
    \midrule
    Teacher$_1$
    & 37.09 & 64.17 & 37.85 & 41.3M & 12.7G \\
    Teacher$_2$
    & 49.61 & 77.31 & 51.35 & 41.3M & 12.7G \\
    Teacher$_3$
    & 54.75 & 83.06 & 59.92 & 41.3M & 12.7G \\
    Teacher$_4$
    & 38.71 & 60.87 & 42.25 & 41.3M & 12.7G \\
    Ensemble
    & 45.04 & 71.35 & 47.84 & 165.2M & 50.8G \\
    \midrule
    Raw
    & 44.94 & 73.52 & 47.16 & 41.3M & 12.7G \\
    Student$_{\text{slim}}$
    & 48.11 (\Blue{+3.17}) & 76.48 (\Blue{+2.96}) & 50.79 (\Blue{+3.63}) & 41.8M & 12.7G \\
    \bottomrule
\end{tabular}}
\label{table:result-4t}
\end{table}

\begin{figure*}[!t]%
\centering%
\subfloat[AP learning curves on VOC dataset]{%
    \includegraphics[scale=1.0]{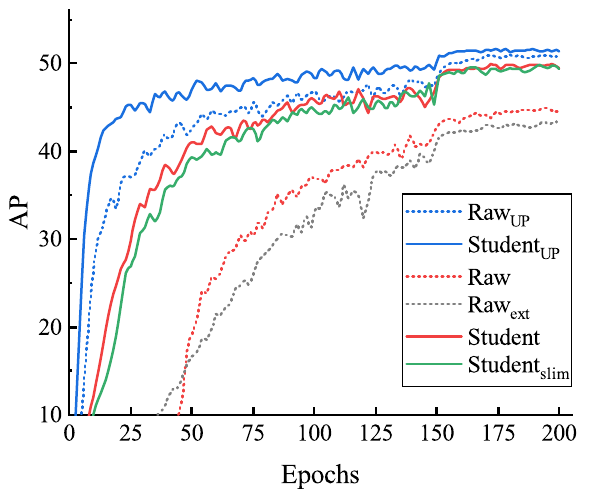}%
    \label{fig:curv-a}%
}%
\subfloat[AP learning curves on COCO dataset]{%
    \includegraphics[scale=1.0]{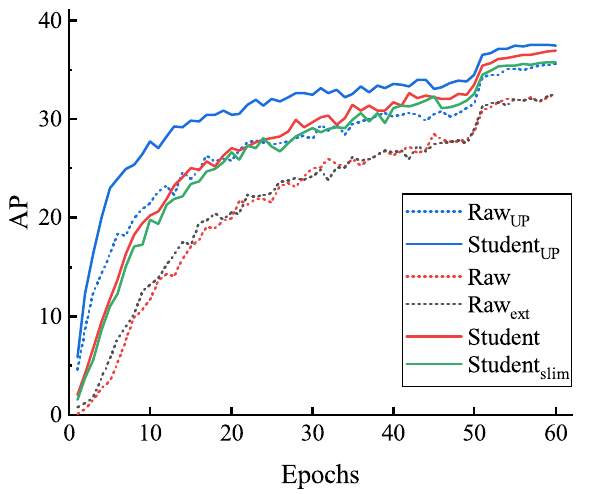}%
    \label{fig:curv-b}%
}%
\subfloat[Ablation study on VOC dataset]{%
    \includegraphics[scale=1.0]{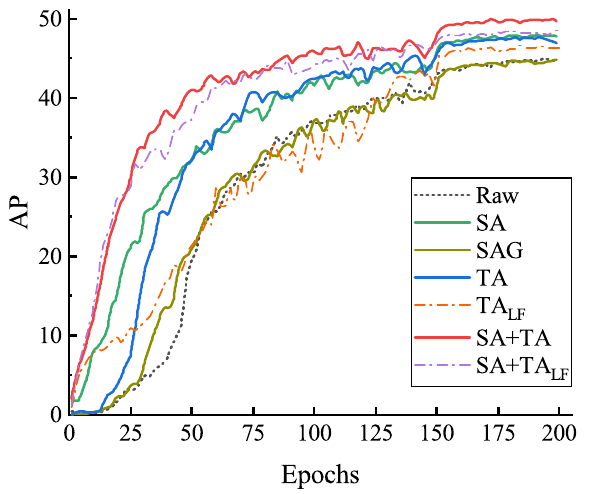}%
    \label{fig:curv-c}%
}%
\caption{Average precision (COCO style) learning curves of our proposed method and baseline settings evaluated on PASCAL VOC and COCO datasets. Specifically, on VOC, we train all models for 150 epochs with the learning rate decayed at 100 epochs with relatively smaller image size than~\cite{dai2020up-detr}; on COCO, we train for only 60 epochs with learning rate decayed at 50 epochs. Besides, we plot AP learning curves of ablation study on loss terms and ground truth labels separately on VOC in (c). Here, SA denotes training with only sequence-level amalgamation term, and SAG denotes sequence aggregation method. TA represents training only with task-level amalgamation term, and LF means training without ground truth labels.%
}%
\label{fig:curv}%
\end{figure*}

We evaluate our proposed KA method on VOC and COCO datasets and set up the following comparison settings.
\begin{itemize}
    \item \textit{Raw}: a DETR baseline model. We train the original DETR model from scratch only with ground truth labels.
    \item \textit{Raw$_\text{ext}$}: a DETR baseline model with sequence extension. Unlike the \textit{raw} setting, we extend the student sequence according to our proposed method to control the influence of sequence extension, which slightly increases the computational cost and the number of parameters.
    \item \textit{Student}: our target student model. According to our proposed KA method, we train the student with both SA and TA supervision.
    \item \textit{Student$_\text{slim}$}: our target student model with sequence compression. We further utilize the sequence compression algorithm to reduce the FLOPs to almost the same as the \textit{raw} setting.
\end{itemize}
Additionally, we also test our proposed method in the circumstance that the student has been unsupervised pre-trained according to~\cite{dai2020up-detr}, which significantly boosts up the student performance (denoted by \textit{UP}).

The comparison results are shown in Table~\ref{table:result}, where the AP, AP$_{50}$ and AP$_{75}$ are demonstrated as the model performance. We present the number of parameters and FLOPs of each method in addition. Among these settings, \textit{ensemble} denotes the ensemble of all the teachers, which is evaluated with all the categories of the corresponding dataset (same as the student). The \textit{blue} number indicates the improvement of AP compared to the raw setting.

From these results, we discover that the student significantly outperforms the baseline setting (increases about 5 percentage points on VOC and 4 percentage points on COCO). Particularly, for the VOC dataset, the student is even superior to the ensemble teacher (with 1.05 percentage points). Moreover, the student only suffers from slight performance degradation with the sequence compression. It is worth noting that the computational costs brought by the sequence extension have not contributed to the final result (decrease about 1.45 percentage points). Consistently, our method draws the same conclusion on COCO.

The AP learning curves of our method compared with the baseline settings on both VOC and COCO datasets are shown respectively in Figure~\ref{fig:curv-a} and~\ref{fig:curv-b}. As we train the student with relatively small image size and insufficient training data (about 16.5k images), Figure~\ref{fig:curv-a} shows that the student converges rapidly with our knowledge amalgamation method and finally outperforms the baseline setting even without the time-costing pre-training procedure. Also, initialized from unsupervised pre-training, the student converges amazingly fast. Figure~\ref{fig:curv-b} illustrates the AP learning curves on COCO, which has larger images and sufficient data (about 118k images). Nonetheless, the student obtains AP of 37.08, surpassing the pre-trained baseline by 1.52 percentage points with the help of our method.

We also validate our method in the four-teacher case, shown in Table~\ref{table:result-4t}. We only demonstrate the student result trained with sequence compression. The student can still master the heterogeneous tasks trained with our KA method (with 3.17 percentage points better than the baseline and 3.07 superior to the ensemble teacher).

\begin{figure*}[!t]%
\centering%
\setlength{\tabcolsep}{0pt}%
\begin{tabular}{*{6}{l}}%
    \includegraphics[width=3cm]{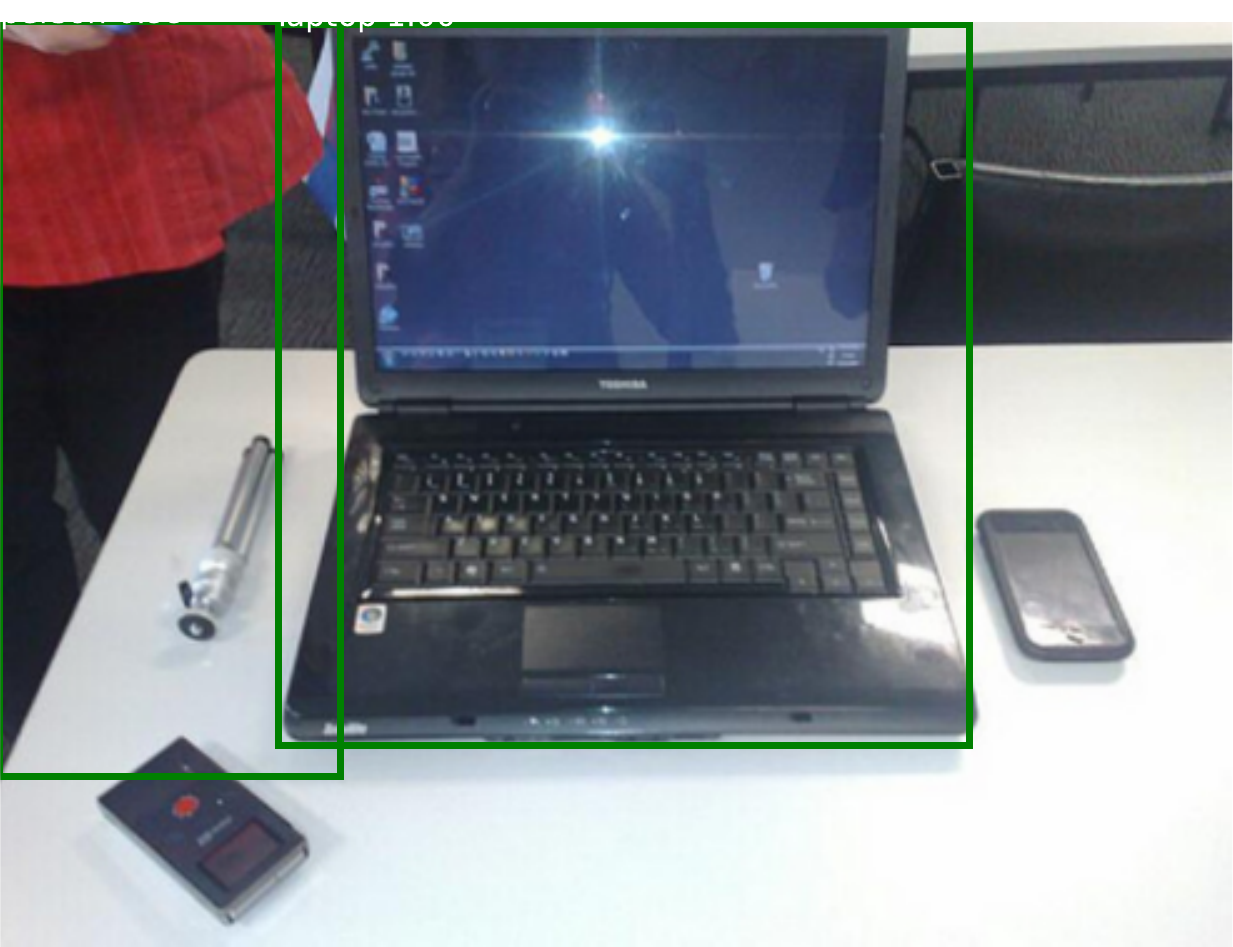}%
    & \includegraphics[width=3cm]{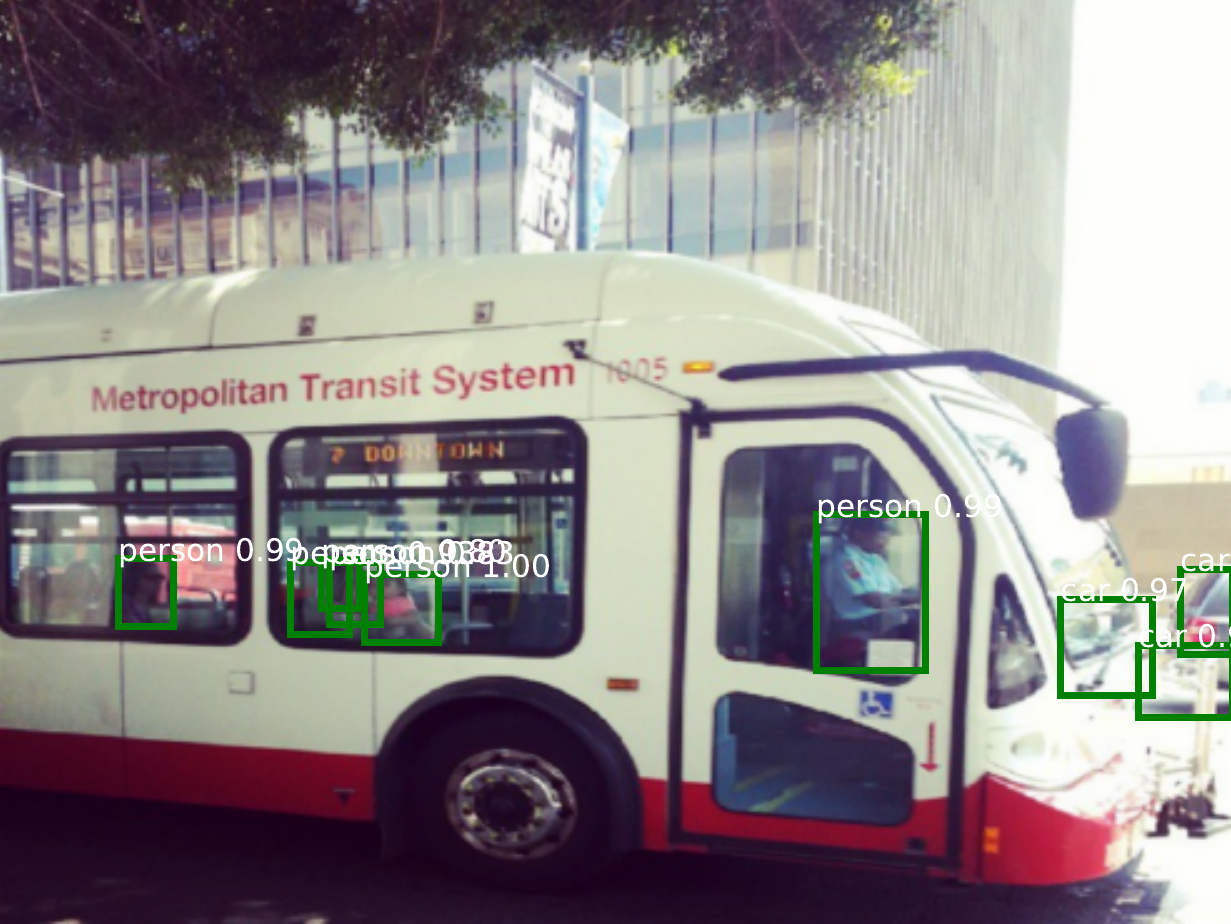}%
    & \includegraphics[width=3cm]{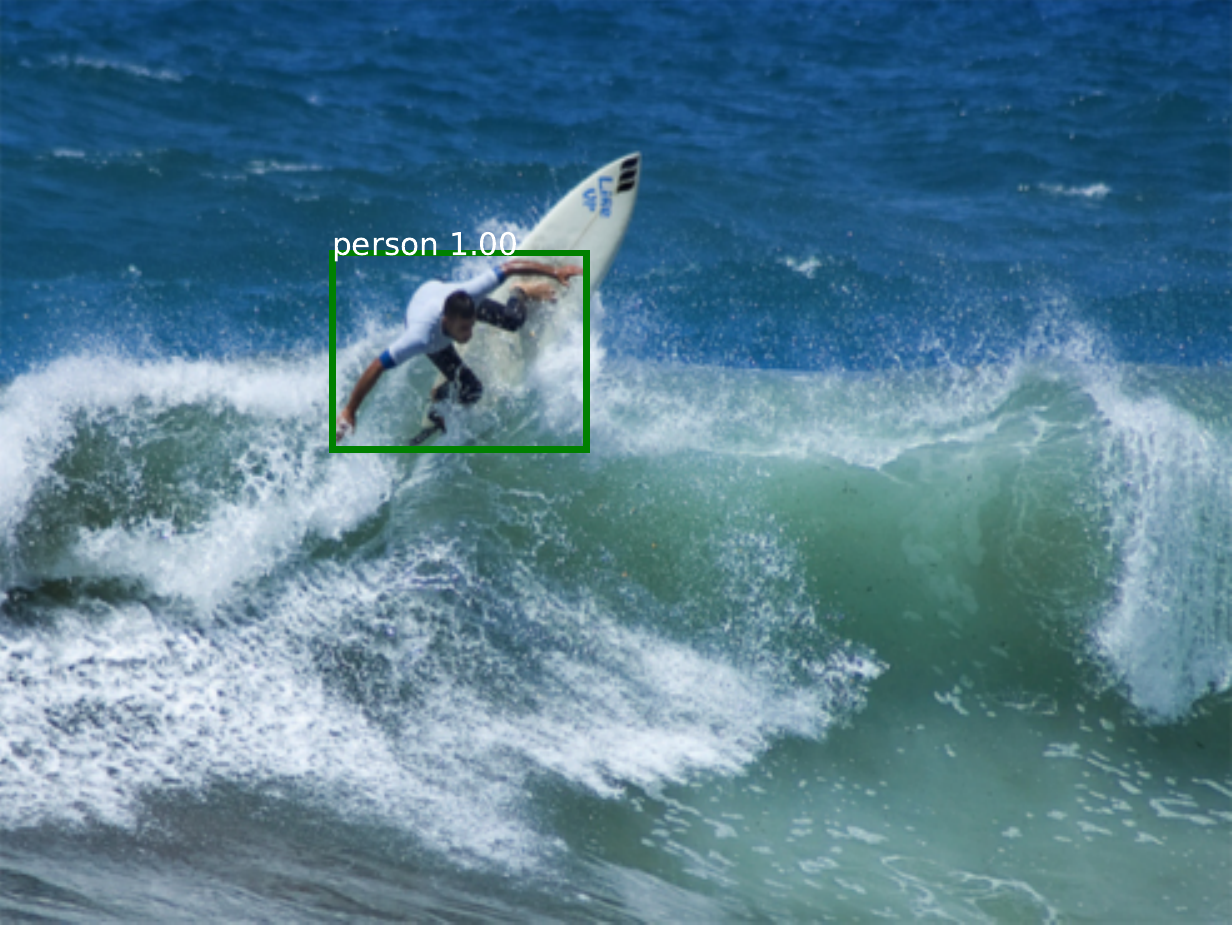}%
    & \includegraphics[width=3cm]{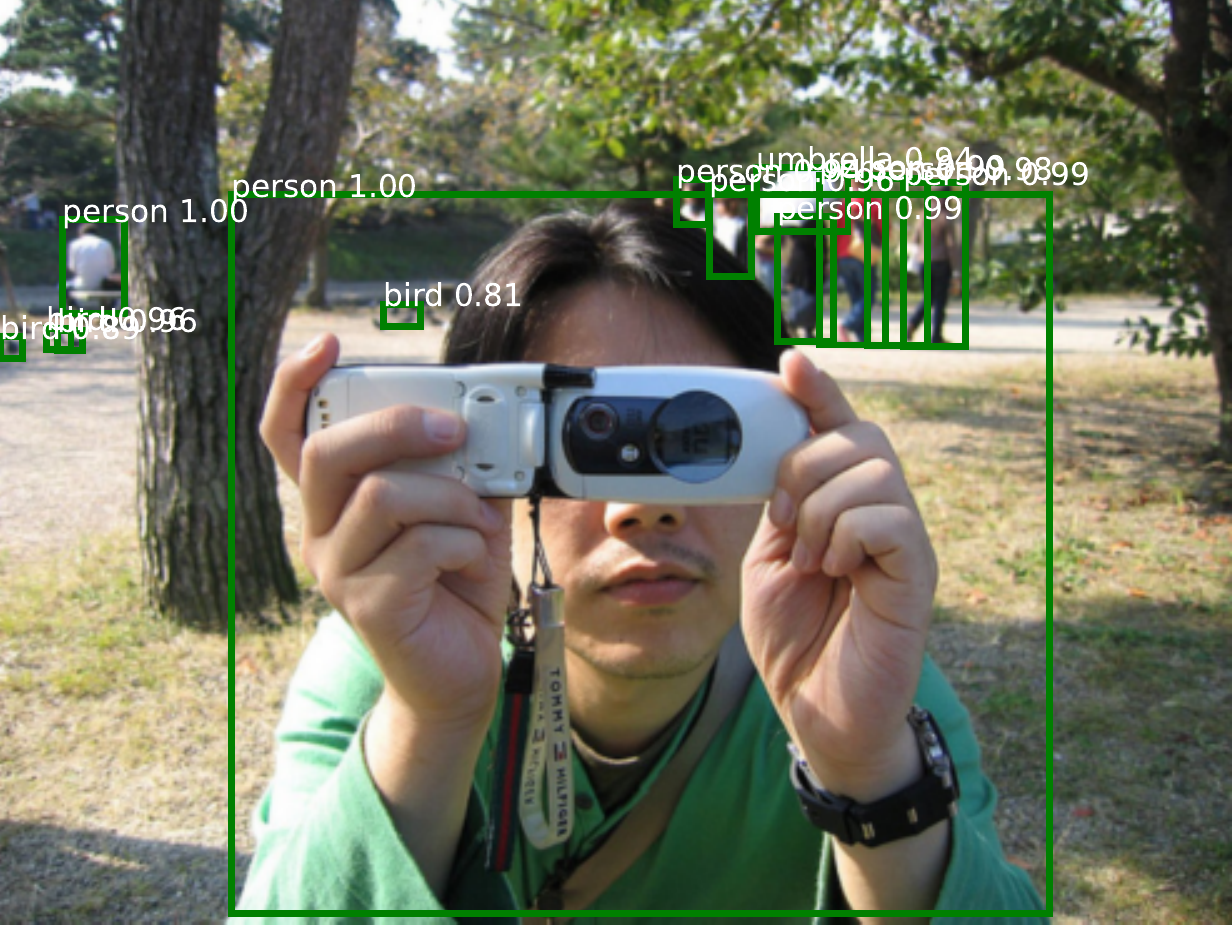}%
    & \includegraphics[width=3cm]{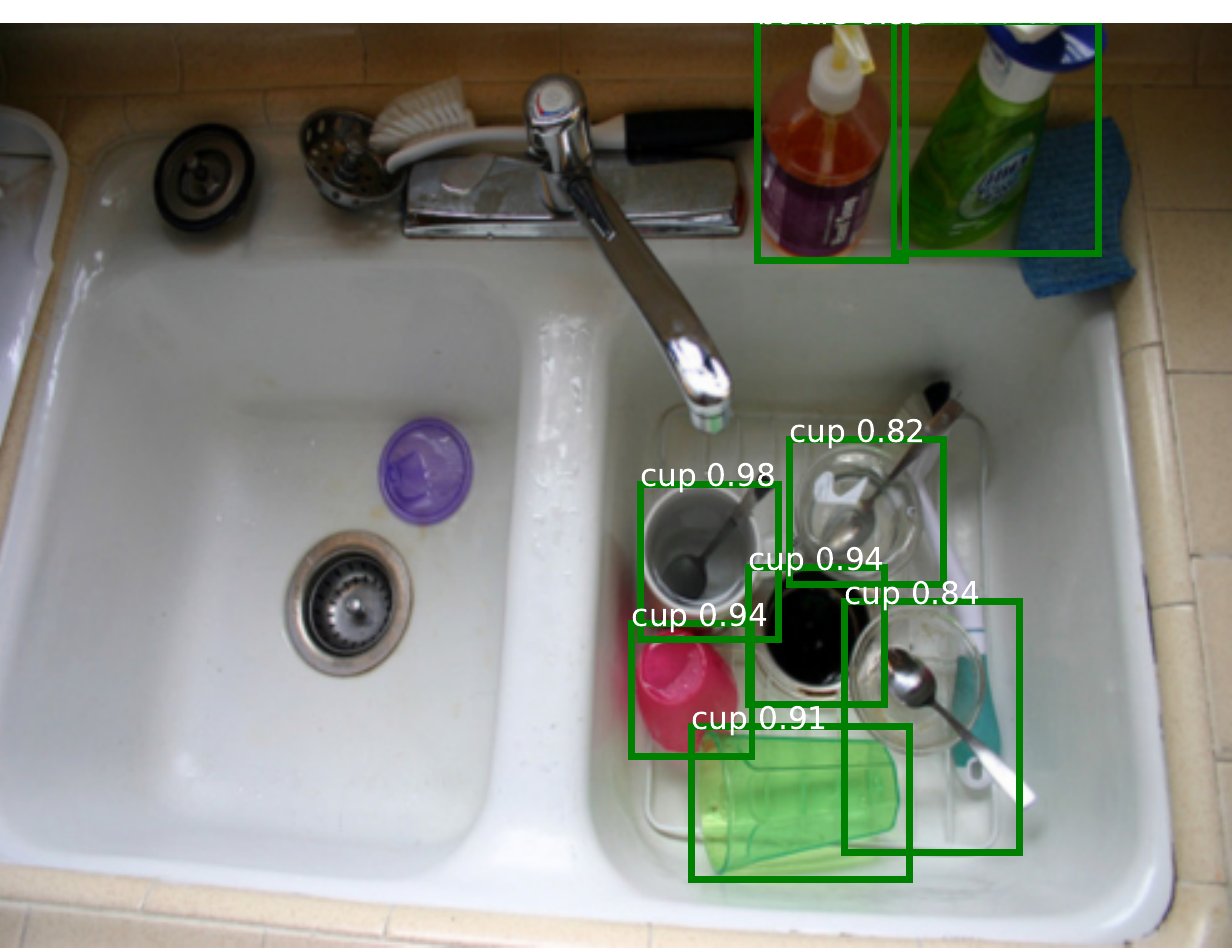}%
    & \includegraphics[width=3cm]{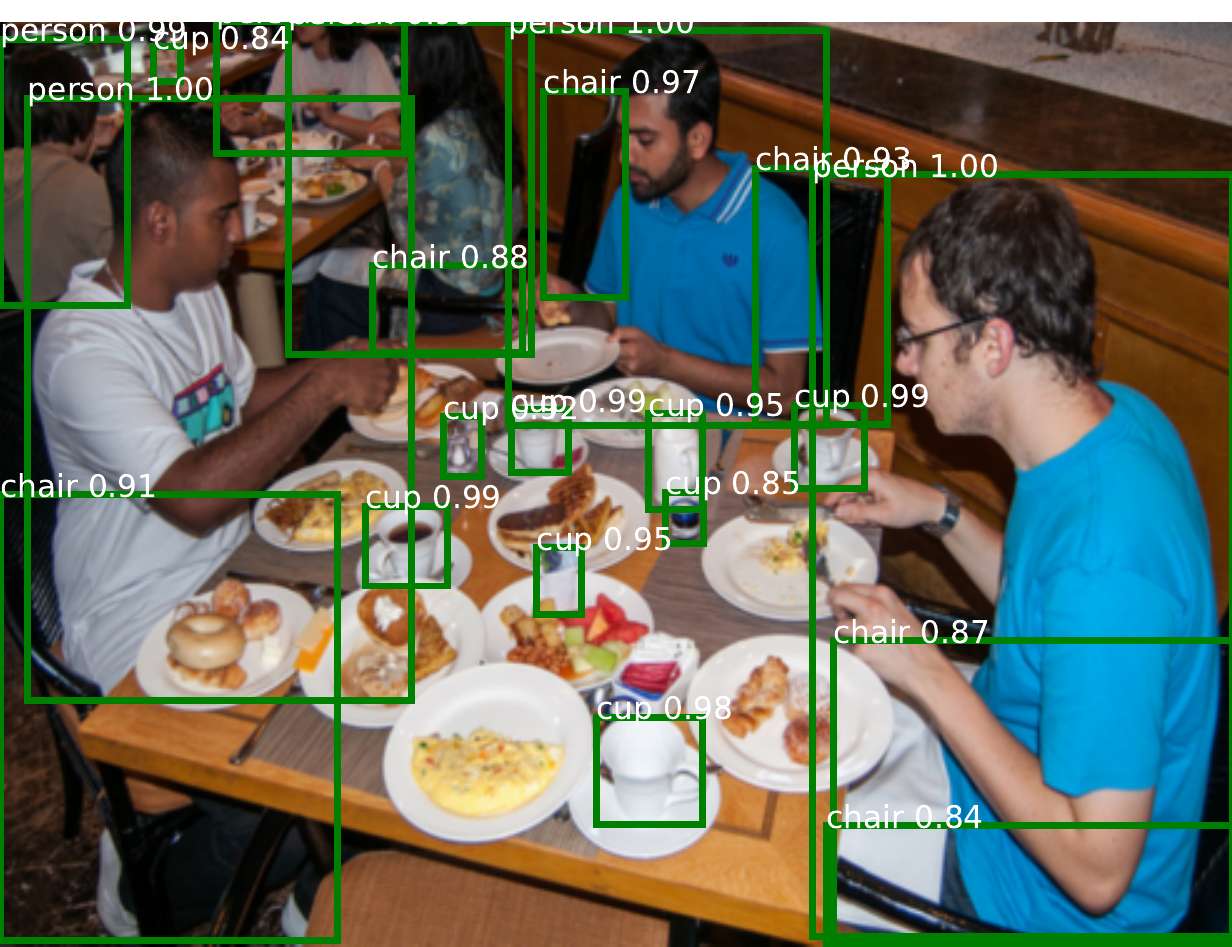} \\%
    \includegraphics[width=3cm]{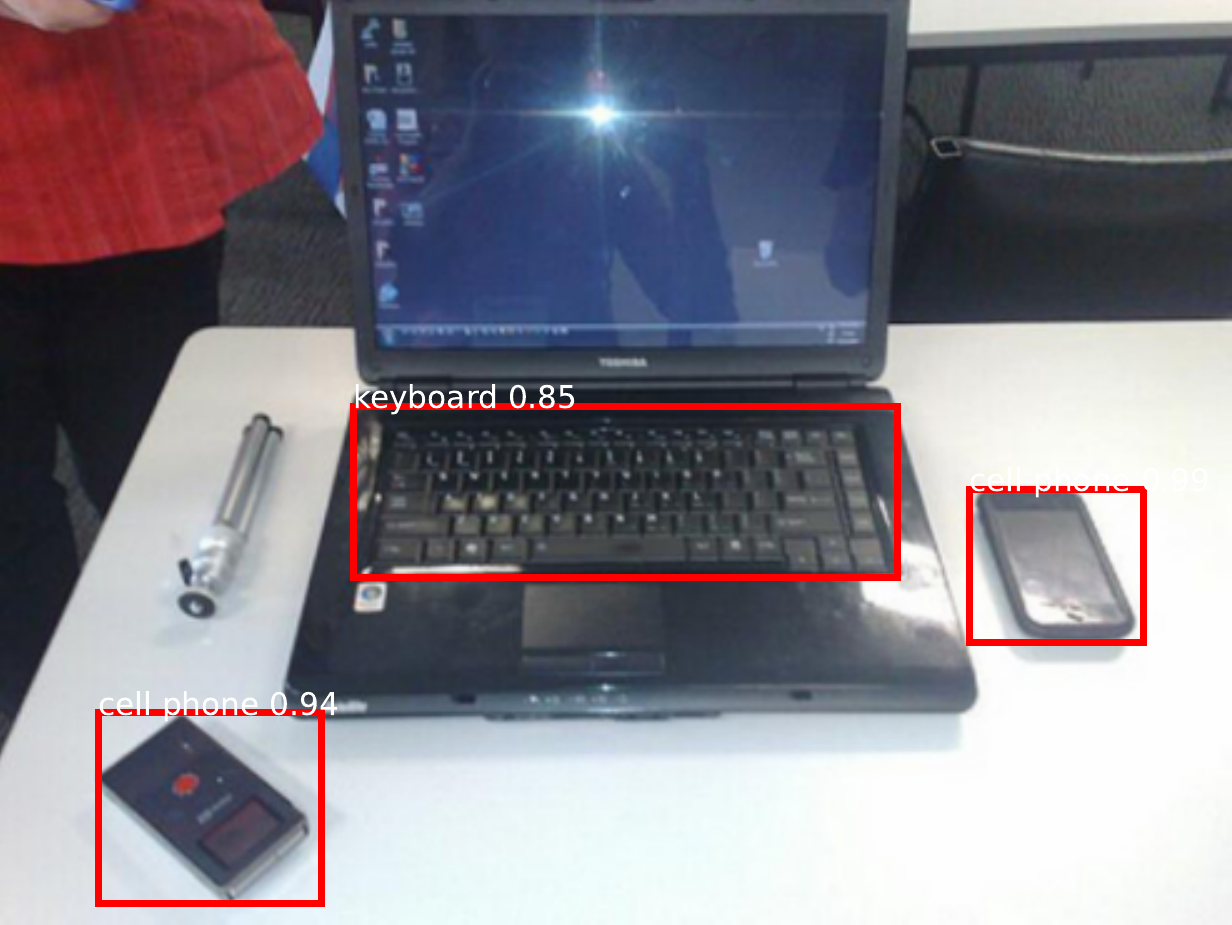}%
    & \includegraphics[width=3cm]{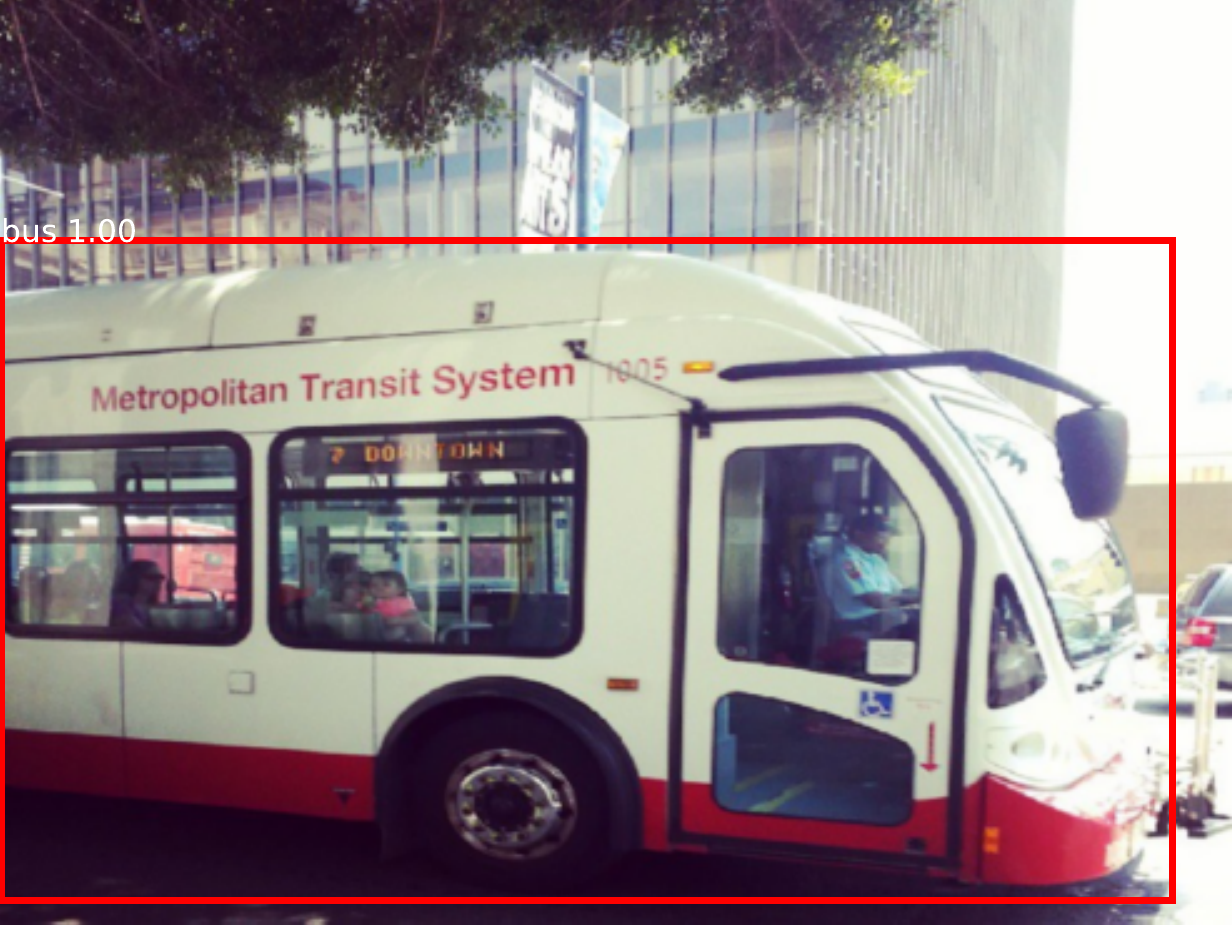}%
    & \includegraphics[width=3cm]{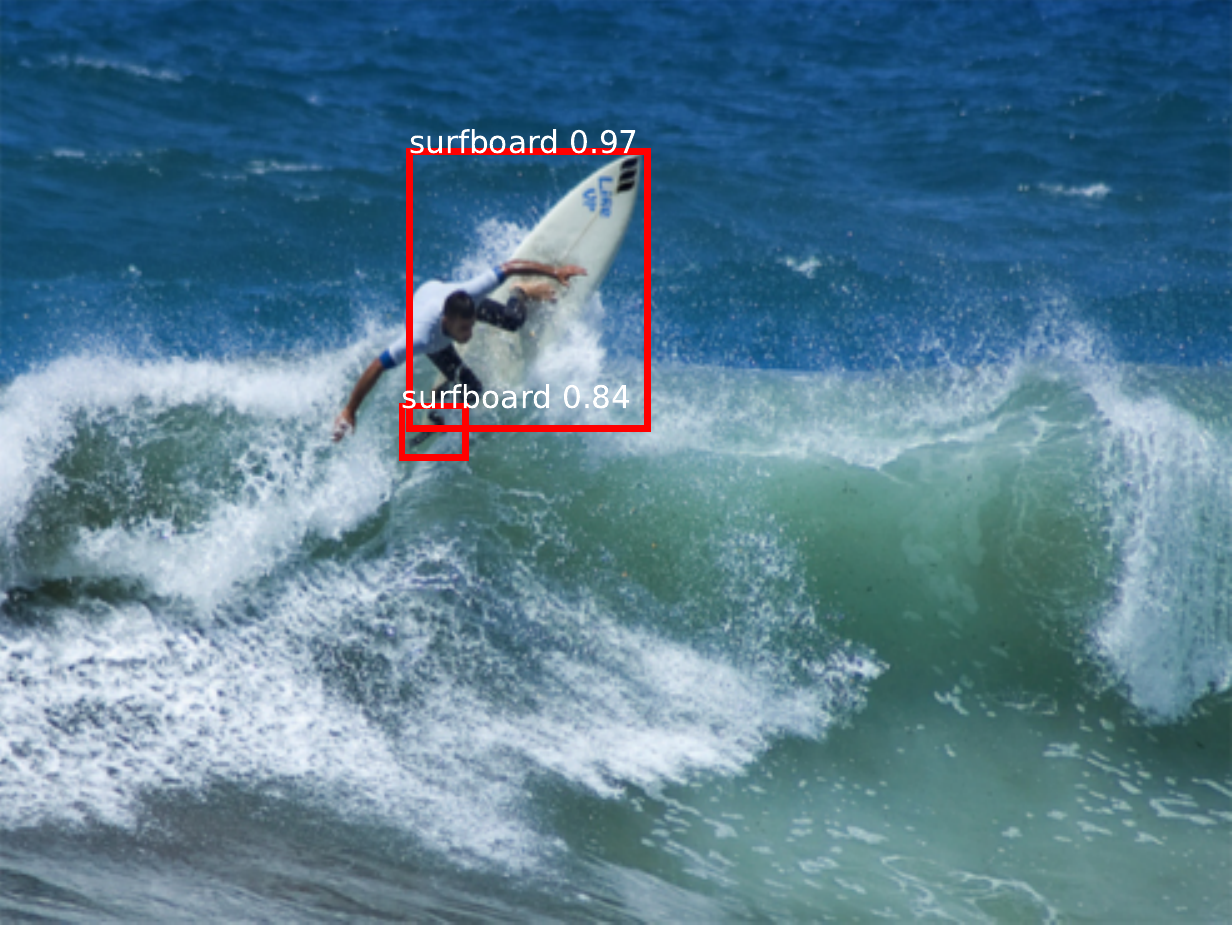}%
    & \includegraphics[width=3cm]{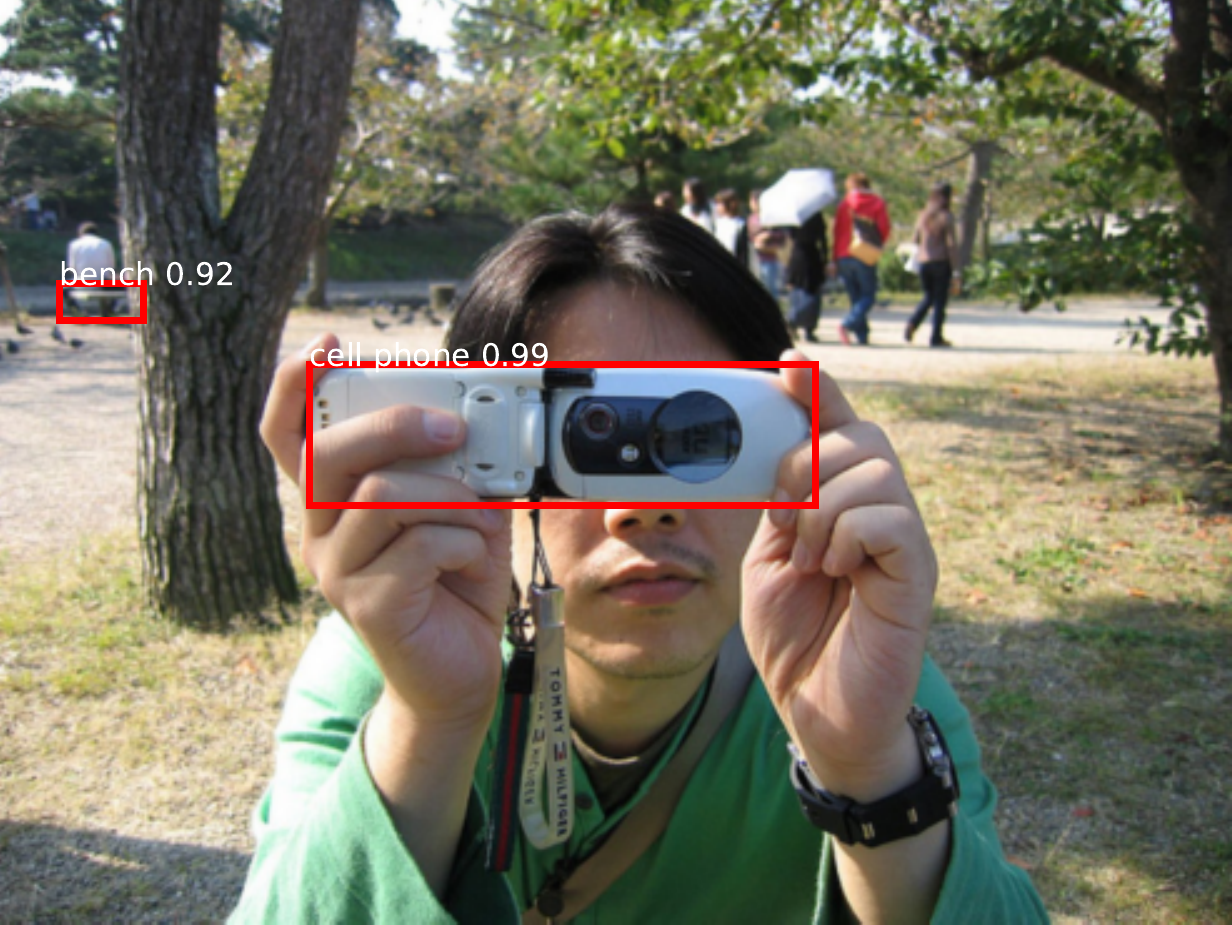}%
    & \includegraphics[width=3cm]{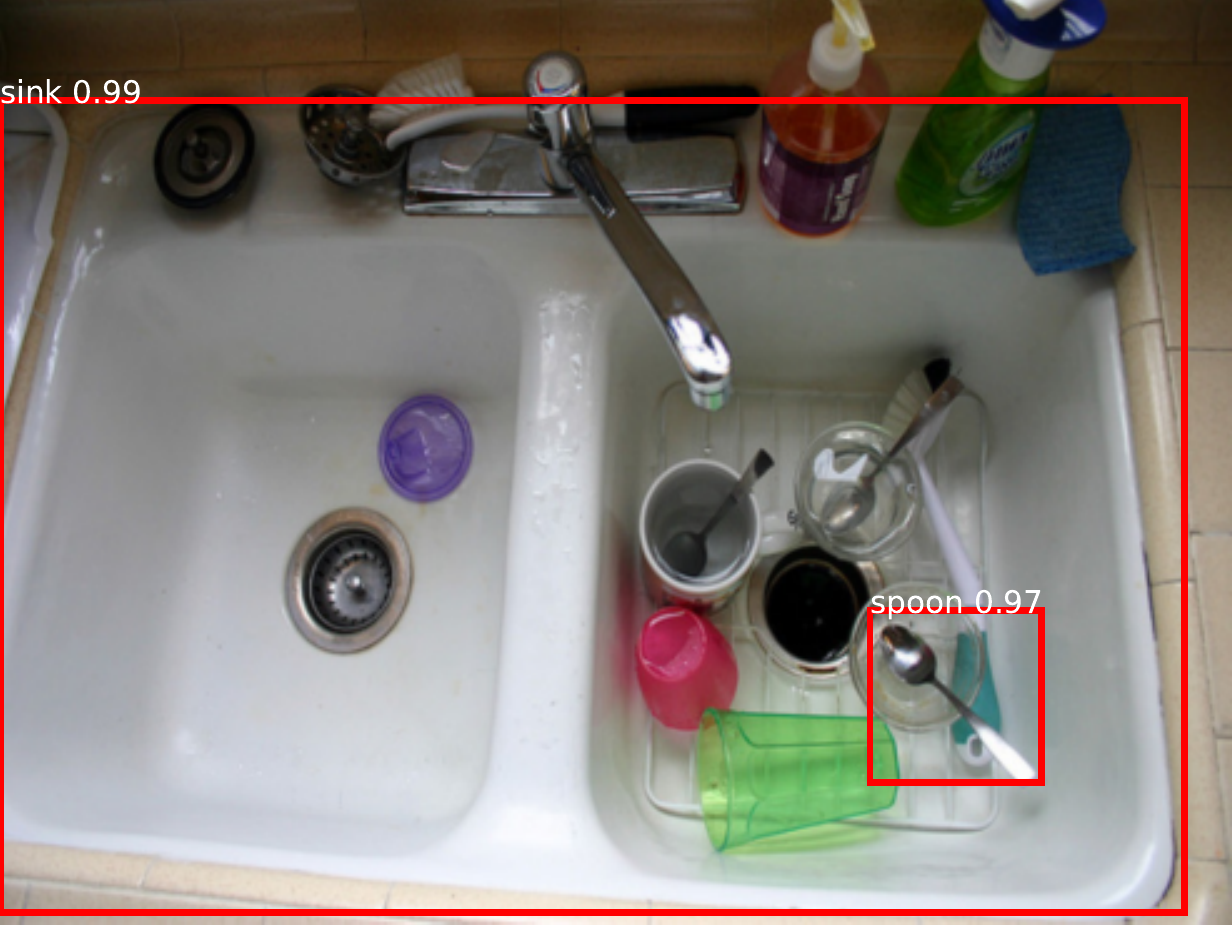}%
    & \includegraphics[width=3cm]{fig/detection/t2-6.pdf} \\%
    \includegraphics[width=3cm]{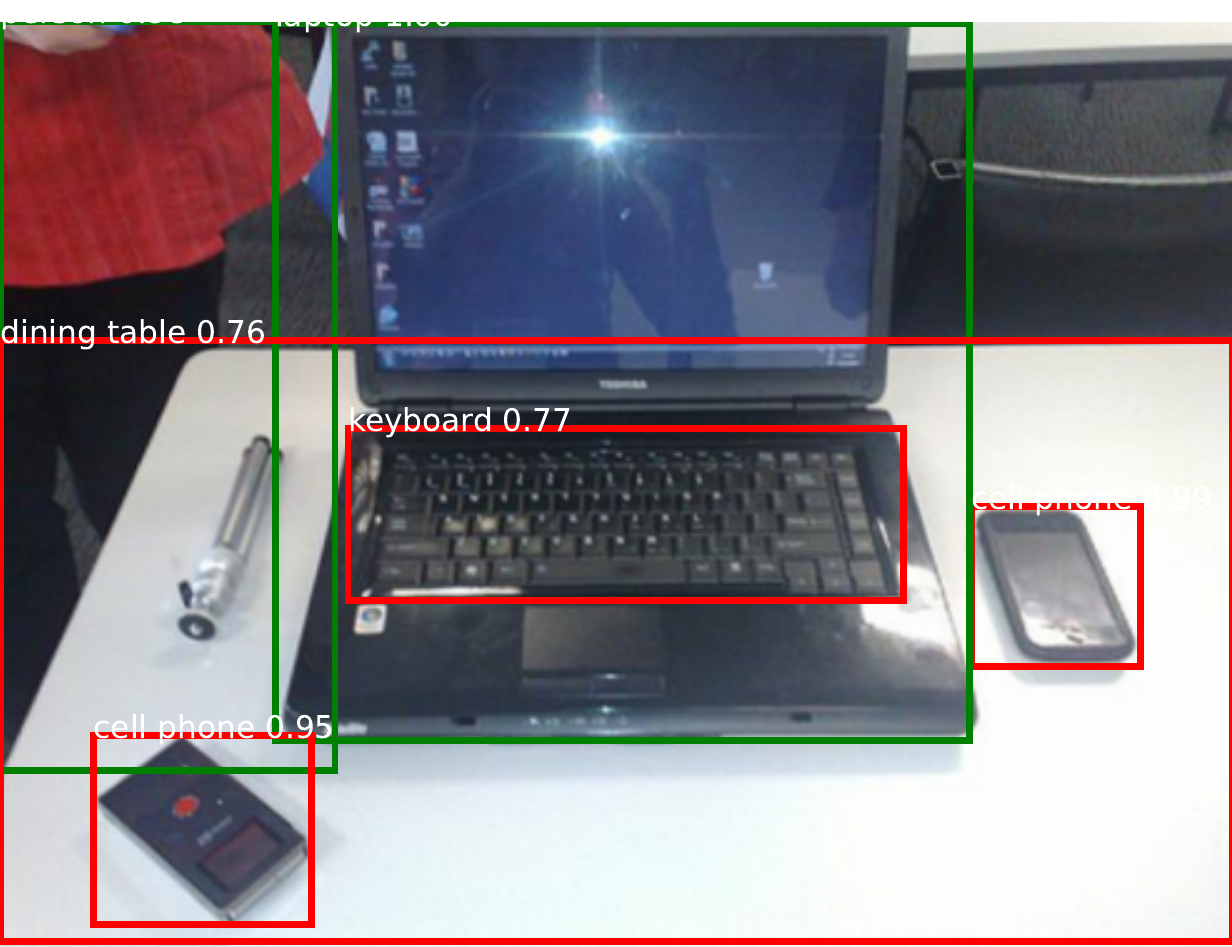}%
    & \includegraphics[width=3cm]{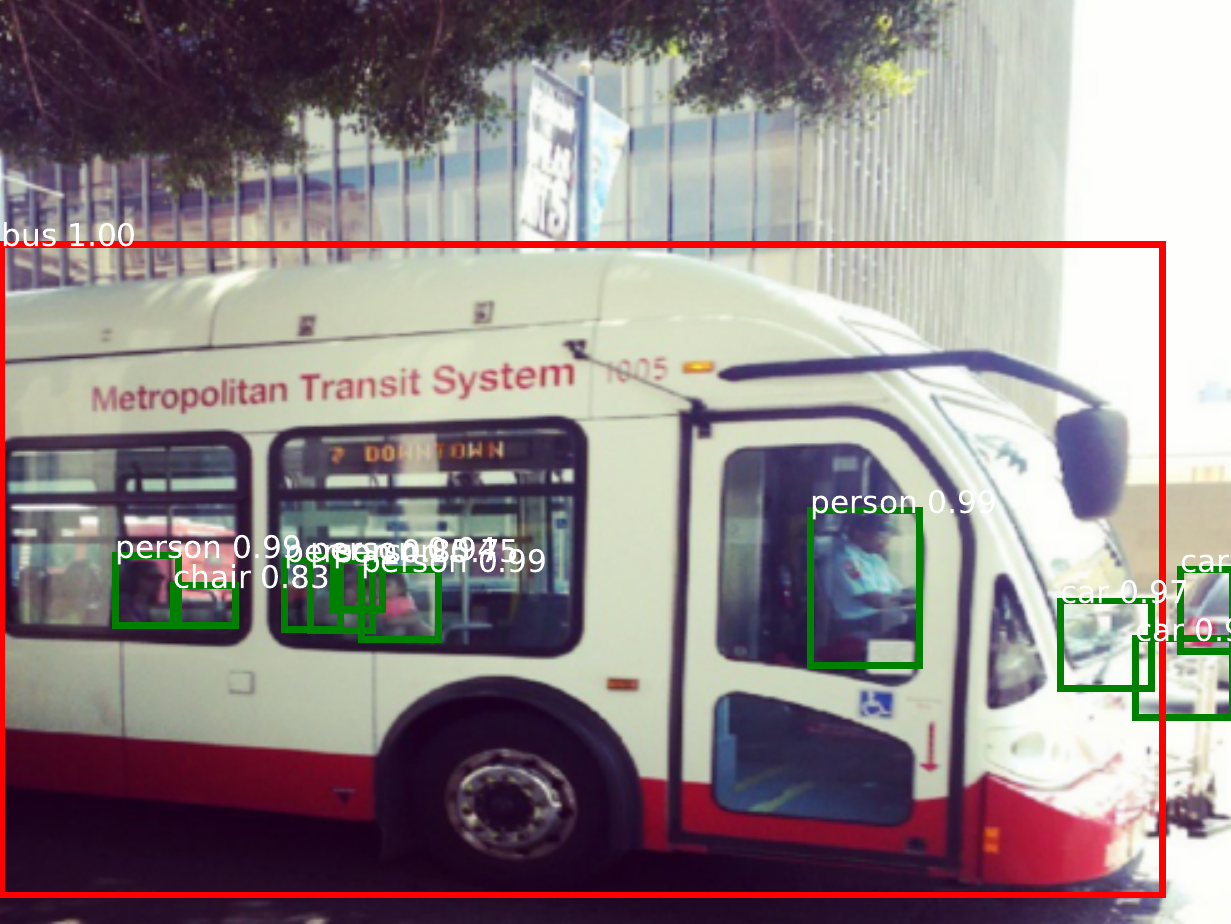}%
    & \includegraphics[width=3cm]{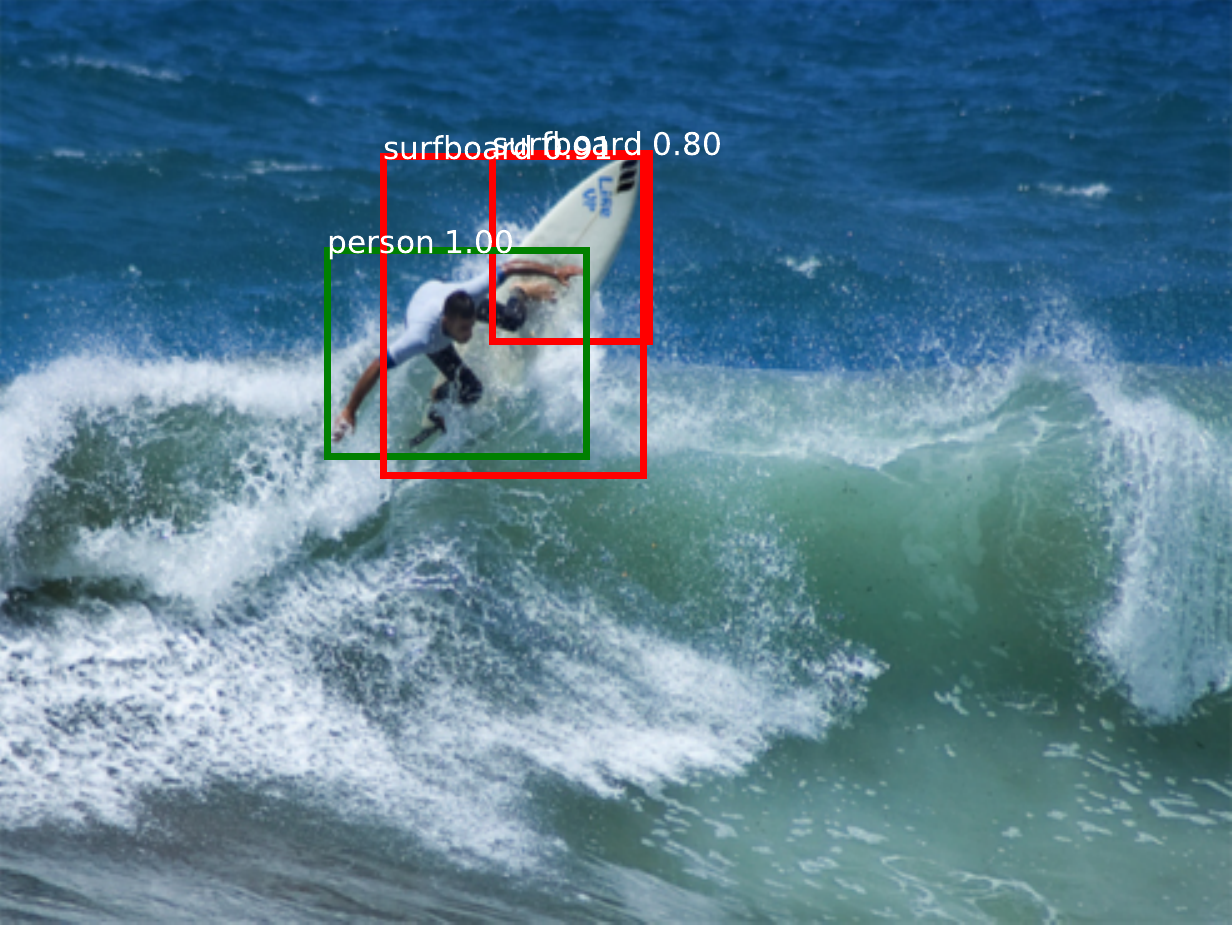}%
    & \includegraphics[width=3cm]{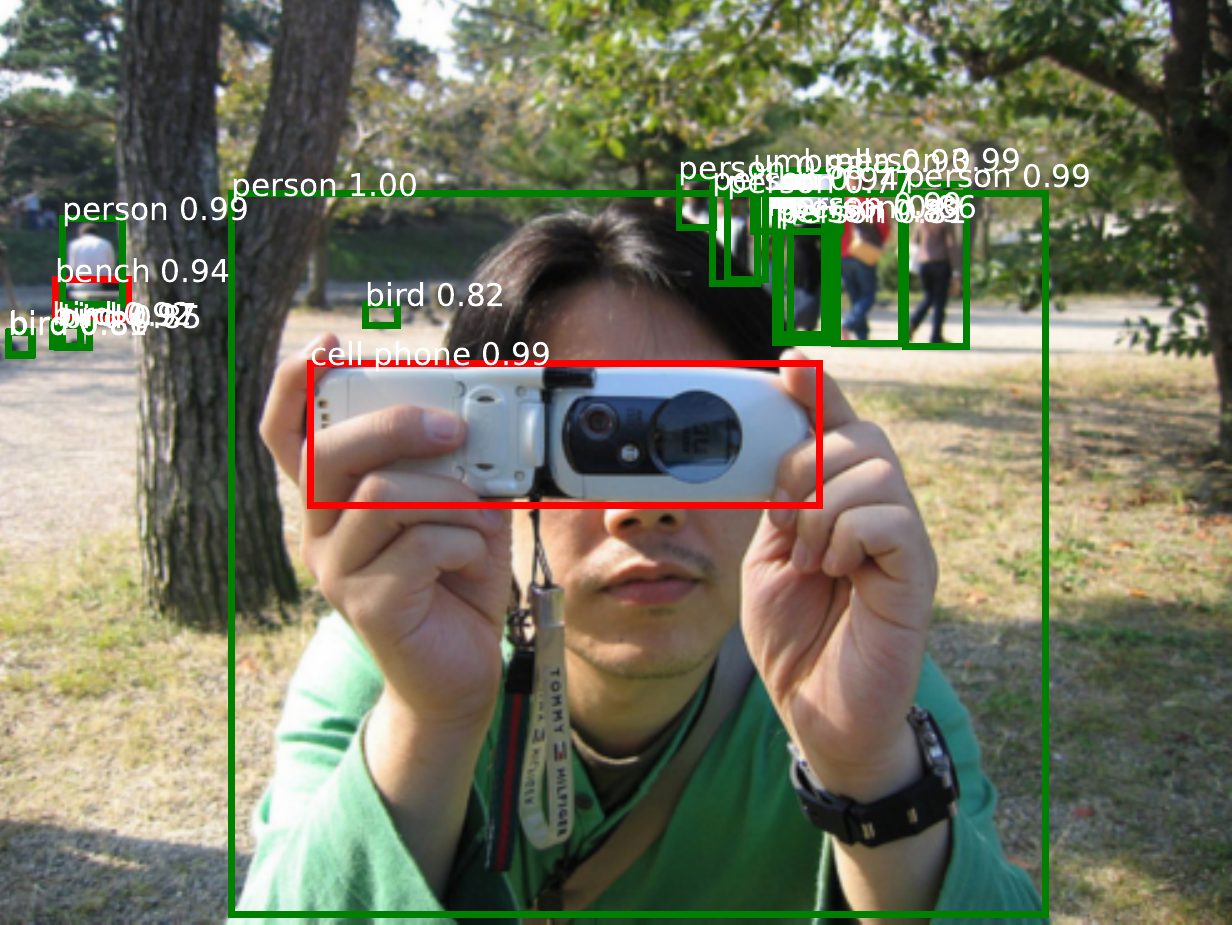}%
    & \includegraphics[width=3cm]{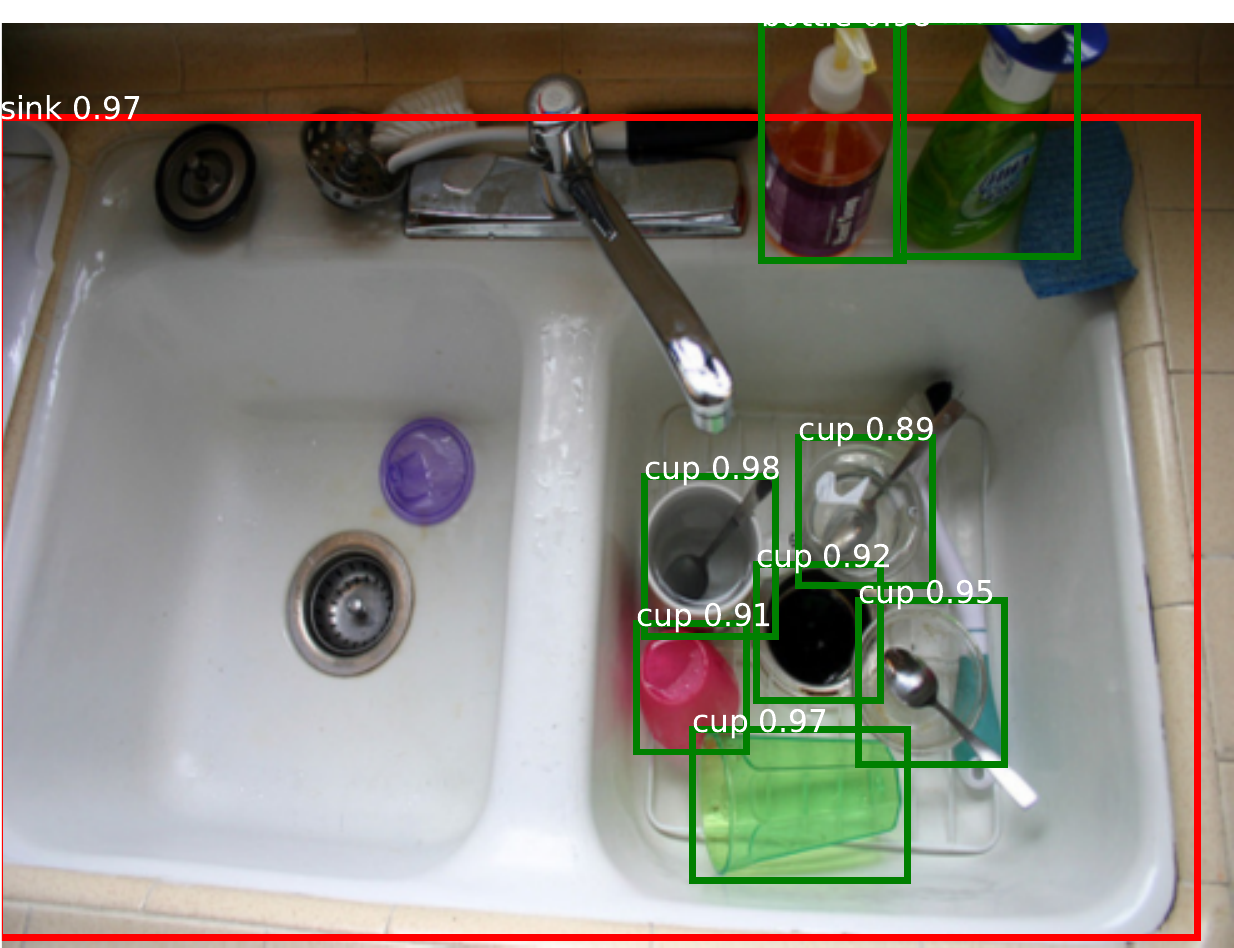}%
    & \includegraphics[width=3cm]{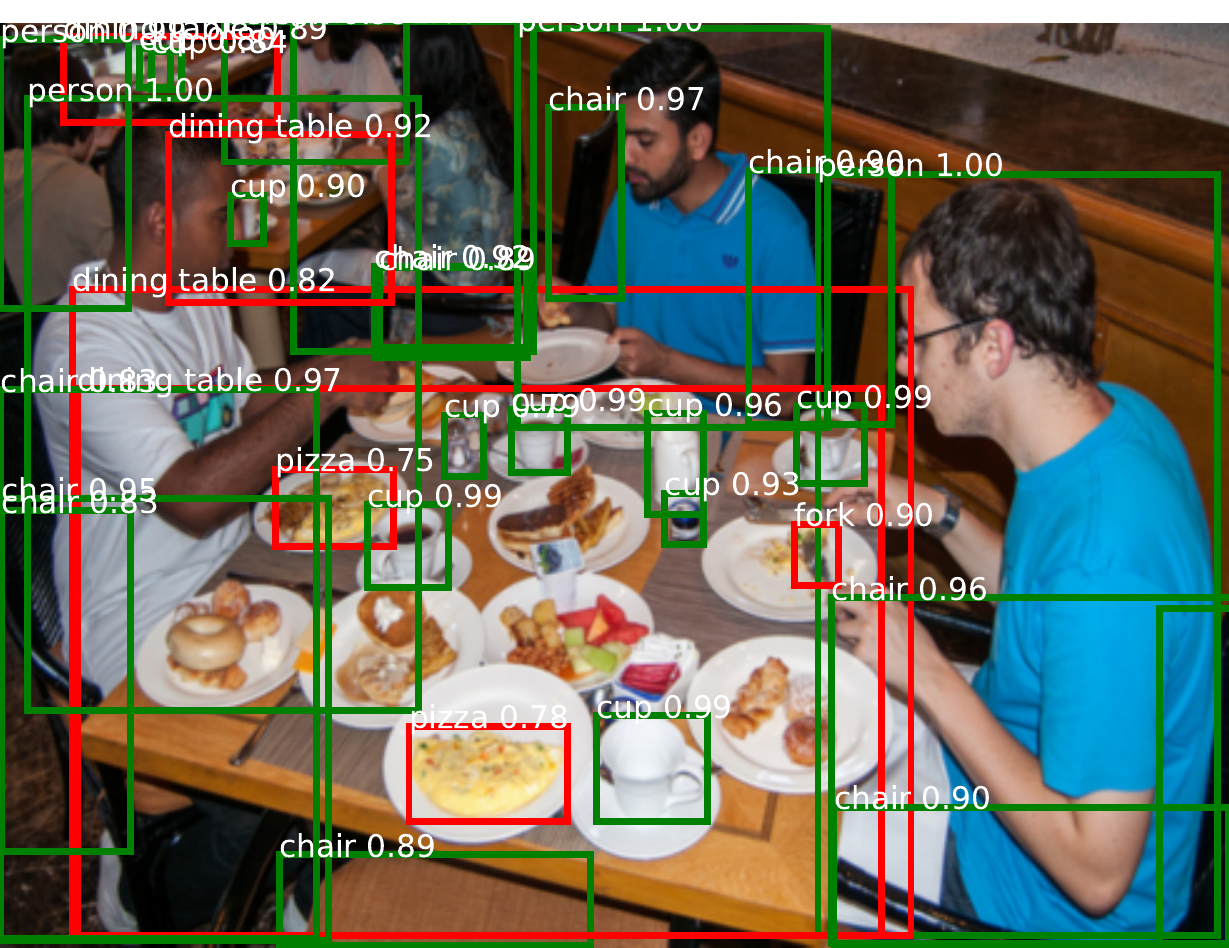} \\%
\end{tabular}%
\caption{Detection results on randomly sampled images from the validation dataset of COCO. The first two rows are detection results of two teachers respectively (with \textit{green}, and \textit{red} bounding boxes) and the final row contains the detection results of the student (with both green and red bounding boxes).}%
\label{fig:vis-det}%
\end{figure*}

\subsection{Ablation Study}
Now, we are going deep into our proposed method to determine the functions of loss terms and ground truth labels. Moreover, we compare the proposed SA method with the sequence aggregation approach widely adopted in previous KA methods for the CNNs.

\subsubsection{Ablation of Loss Terms}
\begin{table}[!t]
\centering
\setlength{\tabcolsep}{4pt}
\caption{Ablation Study of Loss Terms on PASCAL VOC}
\resizebox{\linewidth}{!}{\begin{tabular}{ll *{3}{r} c}
    \toprule
    \multicolumn{1}{l}{\textbf{Settings}}
    & \multicolumn{1}{c}{\textbf{Loss}}
    & \multicolumn{1}{c}{\textbf{AP}}
    & \multicolumn{1}{c}{\textbf{AP$_{50}$}}
    & \multicolumn{1}{c}{\textbf{AP$_{75}$}}
    & \multicolumn{1}{c}{\textbf{Training Time}} \\
    \midrule
    \multirow{2}{*}{\textbf{Sequence}}
    & SA & 47.97 (\Blue{+3.03}) & 76.09 & 50.45 & 34h \\
    & SAG & 44.85 (\Blue{-0.09}) & 73.73 & 46.91 & 34h \\
    \midrule
    \multirow{2}{*}{\textbf{Task}}
    & TA & 47.69 (\Blue{+2.75}) & 77.07 & 50.30 & 39h \\
    & TA$_\text{LF}$ & 46.62 (\Blue{+1.68}) & 74.20 & 48.84 & 38h \\
    \midrule
    \multirow{2}{*}{\textbf{Both}}
    & SA+TA & 49.94 (\Blue{+5.00}) & 78.91 & 52.82 & 41h \\
    & SA+TA$_\text{LF}$ & 48.54 (\Blue{+3.60}) & 76.36 & 51.92 & 40h \\
    \midrule
    \multicolumn{1}{l}{\textbf{Raw}} & - & 44.94 & 73.52 & 47.16 & 25h \\
    \bottomrule
\end{tabular}}
\label{table:ablation}
\end{table}

To figure out the contribution of SA and TA loss terms individually, we show the ablation results in Table~\ref{table:ablation} with AP performance and training time on VOC dataset. We can observe that both SA and TA have contributed to the final result. The SA term has enhanced the student by 3.03, and the TA term has improved by 2.75 percentage points. To further dig out their mastery, we also plot their learning curves in Figure~\ref{fig:curv-c}. It reveals that the SA term helps the student learn faster at the beginning, while the TA term gradually trains a better student. This can be explained as: (1) the SA term supervises the student's intermediate layers, which is a kind of initialization of transformers as~\cite{jiao2019tinybert} reported. Because of this, the initialized student converges faster than the baseline, which is trained from scratch. (2) TA term directly supervises the student with the soften label which acts like soft-label regularization and eventually gets better performance than the baseline setting. With the cooperation of both SA and TA terms, the student learns heterogeneous tasks from teachers rapidly.

\subsubsection{Ablation of Ground Truth Labels}
\label{exp:GT}
Object detection is a relatively more challenging task than image classification, in which most previous knowledge distillation approaches utilize ground truth labels (designed for CNNs). In contrast, previous KA methods for classification are mostly label-free approaches. To further investigate whether the student can learn the amalgamated knowledge without ground truth labels, we report the performance and learning curves of label-free settings in Table~\ref{table:ablation} and Figure~\ref{fig:curv-c} with notation \textit{LF}~(label-free). With both SA and TA supervision, the student still maintains fairly competitive performance (suffers from only 1.40 AP drop). Nevertheless, without SA as initialization to the student, it learns much slower. As a result, the ground truth labels further help train students more steadily, especially without intermediate supervision.

\begin{figure}[!t]%
\centering%
\subfloat[Redundancy]{%
    \includegraphics[width=0.49\linewidth]{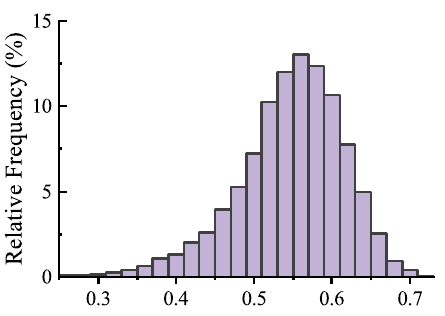}%
    \label{fig:seq-redundancy}%
}%
\subfloat[Methods]{%
    \includegraphics[width=0.49\linewidth]{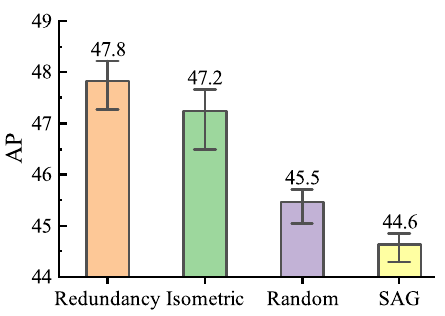}%
    \label{fig:seq-compression}%
}%
\caption{%
    Figure~(a) illustrates the redundancy distribution of extended student sequences over the VOC dataset. Here, we visualized the extended student sequence output from the linear projection module. Figure~(b) compares different sequence compression methods with $50\%$ keep rate, including the previous sequence aggregation approach. We repeat experiments five times and plot their performance with an error bar (the best and worst cases).%
}%
\label{fig:seq-analysis}%
\end{figure}

\subsubsection{Comparison with Sequence Aggregation}
\label{exp:ablation-agg}

We evaluate our proposed sequence-level amalgamation with compression (for a fair comparison) and the SAG approach introduced in Section~\ref{sec:seq_amg_analysis}. As shown in Table~\ref{table:ablation}, the sequence aggregation approach suffers from significant performance degradation (even worse than the baseline setting). Moreover, the learning curves in Figure~\ref{fig:curv-c} demonstrate that, throughout the whole training procedure, the student barely benefits from the SAG supervision. This reflects that transformers are somehow more sensitive to noisy supervision (introduced by the projection weights) than CNNs, as the MHA module is good at modeling the relationship between individual vision tokens, whereas the convolution kernels focus on the neighbor features which tolerant the disturbance of noises.

\subsection{Vision Sequence Analysis}
In this section, we dig into the vision sequence to further analyze the redundancy of vision tokens and compare different sequence compression strategies.

\subsubsection{Redundancy Analysis}
\label{exp:redundancy}

We present the redundancy distribution of vision tokens in Figure~\ref{fig:seq-redundancy}. The student's extended sequences output from linear projection models (as shown in Figure~\ref{fig:seq-exten}) are extracted to calculate redundancy defined in Equation~\ref{eq:redundancy}. As we can see, over half of the vision tokens have $R > 0.5$, which makes it possible to train a relatively compact student with limited performance degradation.

\subsubsection{Comparison of Different Compression Strategies}
\label{exp:compression}

We compare different compression strategies mentioned in Section~\ref{sec:seq-compression}, such as isometric down-sampling, random down-sampling, and redundancy-based compression in Figure~\ref{fig:seq-compression}. We repeat the experiments five times with $50\%$ compression rate. It illustrates that the redundancy-based sequence compression method performs better than the isometric down-sampling (by 0.6 percentage points) and significantly outperforms random down-sampling and SAG methods.

\subsection{Comparison with Previous CNNs}
We list the performance of our method and several widely adopted CNN-based object detection methods on COCO in Table~\ref{table:sota}. Although the student is trained for just 60 epochs, its performance is still attractive. And trained with more epochs, such as 150, the student achieves competitive performance. Nevertheless, we must clarify that our goal is \textit{not} to surpass current object detection methods on each benchmark but rather explore a knowledge amalgamation method for transferring knowledge as much as possible from heterogeneous teachers.
\begin{table}[!t]
\centering
\setlength{\tabcolsep}{4pt}
\caption{Comparison with CNNs}
\begin{threeparttable}
\begin{tabular}{*{4}{l}}
    \toprule
    \textbf{Model} & \textbf{Backbone} & \textbf{Epochs} & \textbf{AP} \\
    \midrule
    Mask R-CNN \tnote{$\dagger$} & R101-FPN & - & 38.2 \\
    Grid R-CNN \tnote{$\dagger$} & R101-FPN & - & 41.5 \\
    Double-head R-CNN \tnote{$\dagger$} & R101-FPN & - & 41.9 \\
    RetinaNet \tnote{$\dagger$} & R101-FPN & - & 39.1 \\
    FCOS \tnote{$\dagger$} & R101-FPN & - & 41.5 \\
    Faster R-CNN \tnote{$\dagger$} & R50-FPN & 3$\times$ & 40.2 \\
    DETR \tnote{$\dagger$} & R50 & 150 & 39.7 \\
    UP-DETR \tnote{$\dagger$} & R50 & 150 & 40.5 \\
    UP-DETR \tnote{$\dagger$} & R50 & 300 & 42.8 \\
    Ours & R50 & 60 & 37.7 \\
    Ours & R50 & 150 & 41.8 \\
    \bottomrule
\end{tabular}
\begin{tablenotes}
    \item[$\dagger$] Results are reported in~\cite{dai2020up-detr}.
\end{tablenotes}
\end{threeparttable}
\label{table:sota}
\end{table}

\subsection{Visualization}

\subsubsection{Visualization of Detection Result}
We show the detection results of our students on the COCO dataset in Figure~\ref{fig:vis-det}. The first and second rows demonstrate the detection result of two teachers, and the final row shows the results of the student. From these results, we can see that the student has the ability to imitate heterogeneous teachers and detect objects even in crowded spaces.

\subsubsection{Visualization of Intermediate Sequences}
\begin{figure}[!t]%
\centering%
\subfloat[]{%
    \includegraphics[width=0.49\linewidth]{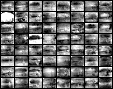}%
    \label{fig:seq-a}%
}\hfil%
\subfloat[]{%
    \includegraphics[width=0.49\linewidth]{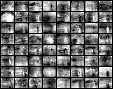}%
    \label{fig:seq-b}%
} \\
\subfloat[]{%
    \includegraphics[width=0.49\linewidth]{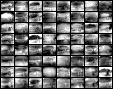}%
    \label{fig:seq-c}%
}\hfil%
\subfloat[]{%
    \includegraphics[width=0.49\linewidth]{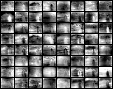}%
    \label{fig:seq-d}%
}%
\caption{%
    Visualization of the intermediate sequence of teachers and the student of input figure shown in Image~\ref{fig:vis-seq-img}. The first row contains the teacher sequence $X_t^1$ and $X_t^2$ respectively in (a) and (b) and the second row contains $X_s^1$ and $X_s^2$ in (c) and (d). We reshape an intermediate sequence to the feature map with shape $w' \times h' \times d$ and show each channel individually in a sub-figure (each channel is a figure with shape $13 \times 10$ pixels).%
}%
\label{fig:vis-seq}%
\end{figure}

We further show the visualization results of the intermediate sequence of teachers and the student. These sequences are reshaped to the original feature map as the output of the CNN backbone and we plot each channel to an individual image (as the channel number of the feature map is 256, there are 256 grayscale images).

\begin{wrapfigure}{r}{0.45\linewidth}%
\centering%
\includegraphics[width=1.0\linewidth]{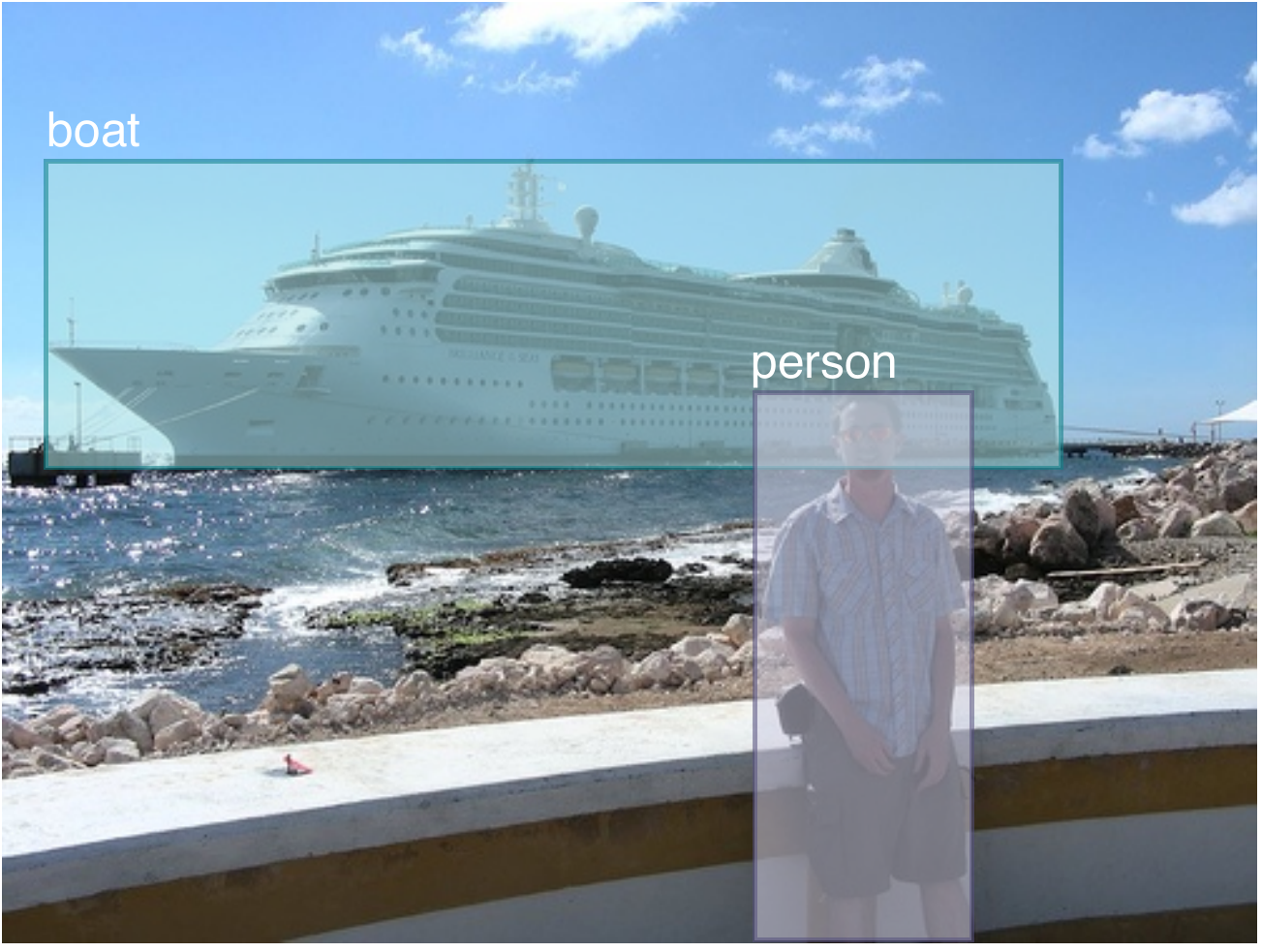}%
\caption{Selected image from VOC dataset that contains a boat (learned by $T_1$) and a person (learned by $T_2$).}%
\label{fig:vis-seq-img}%
\end{wrapfigure}

To illustrate the differences between the two teachers, we randomly pick images with objects these teacher learned separately, as shown in Figure~\ref{fig:vis-seq-img}. The output sequences of the final encoder are extracted and normalized for better visual effect. The intermediate sequences are shown in Figure~\ref{fig:vis-seq}. The first row contains the teacher sequence $X_t^1$ and $X_t^2$.
Similarly, the second row contains the student sequence $X_s^1$ and $X_s^2$. As we can observe, on the one hand, the teacher $T_1$ mainly concentrates on the boat object, and conversely, $T_2$ pays more attention to the person. On the other hand, without cross attention and supervised by those teachers separately, both of the student sequences $X_s^1$ and $X_s^2$, learn the boat and the person object. This shows that the student somehow manages to grasp the amalgamated knowledge during the backward propagation.

\section{Conclusion}
In this paper, we propose to dissolve knowledge amalgamation for transformer-based object detection models.
Our goal is to train a versatile yet lightweight transformer-based student from several well-trained teachers specialized in heterogeneous object detection tasks.
Towards this end, we propose sequence-level amalgamation~(SA) and task-level amalgamation~(TA) methods to transfer knowledge as much as possible.
Particularly, for the characteristic of the transformers, the intermediate layers of the student can be supervised explicitly by concatenated teacher sequences rather than redundantly aggregating them to fixed-size ones as previous KA works for the CNNs. 
Extensive experiments on PASCAL VOC and COCO datasets demonstrate that transformer-based students excel in learning heterogeneous object detection tasks, as they achieve performance on par with or even superior to those of the teachers in their specializations.

\appendices
\newtheorem{theorem}{Theorem}
\newtheorem{lemma}{Lemma}
\newtheorem{property}{Property}

\section{Derivation of Multi-Head Attention}
\label{appendix:1}
In this section, we derivate the matrix form of multi-head attention. Given queries $Q$, keys $K$, and values $V$, the original one head attention mechanism \cite{NIPS2017_attention} is
\begin{equation*}
    att(Q, K, V) = \sigma\left(\frac{Q K^\top}{\sqrt{d_k}}\right)V\text{.}
\end{equation*}
The multi-head attention mechanism is the concatenated outputs of all heads with linear transform
\begin{equation*}
    \text{MHA}(Q, K, V) = \begin{bmatrix}
        H_1 & \cdots & H_h
    \end{bmatrix}
    \begin{bmatrix}
        W_1^O \\
        \vdots \\
        W_h^O
    \end{bmatrix}\text{,}
\end{equation*}
where $W_i^O\in\mathbb{R}^{d_v\times d_{\text{model}}}$ are output projection matrices. And each head is
\begin{equation*}
    \begin{aligned}
        H_i &= att(QW_i^Q, KW_i^K, VW_i^V) \\
            &= \sigma\left(\frac{Q W_i^Q {W_i^K}^\top K^\top}{\sqrt{d_k}}\right) V W_i^V \text{.}
    \end{aligned}
\end{equation*}
Here, $W_i^Q\in\mathbb{R}^{d_{\text{model}\times d_k}}$, $W_i^K\in\mathbb{R}^{d_{\text{model}\times d_k}}$ and $W_i^V\in\mathbb{R}^{d_{\text{model}\times d_v}}$ are projection matrices for each head.

To simplify the notations, let $A_i = Q W_i^Q {W_i^K}^\top K^\top$ be the attention matrix, and $W_i^{VO} = W_i^V W_i^O$ be the combined projection matrix, so the final matrix form of the MHA is
\begin{equation*}
    \text{MHA}(Q, K, V) = \sum_{i=1}^h \sigma\left(\frac{A_i}{\sqrt{d_k}}\right) V W_i^{VO} \text{.}
\end{equation*}

\section{Proof of Transformer Permutation Invariance}
\label{appendix:3}

In this section, we prove that encoder-decoder structured transformers have the property of permutation invariance. It means that given an input sequence $X=(x_1, \ldots, x_n)$, no matter how we permute $X$, the output will not change. Let $P=(p_1, \ldots, p_n)$ denotes a permutation of sequence $(1, \ldots, n)$, where $1 \leq p_i \leq n$, $p_i\neq p_j$ for any $i\neq j$. Then, the permuted sequence is written as $X_P = (x_{p_1}, \ldots, x_{p_n})$. For convenience, permutation can be written in matrix form $\varPhi \in \{0, 1\}^{n\times n} \in \mathcal{P}_n$ where
\begin{equation*}
    \varPhi_{i,j} = \begin{cases}
        1 & \mbox{if } p_i = j \text{,}\\
        0 & \mbox{otherwise} \text{,}
    \end{cases}
\end{equation*}
and $\mathcal{P}_n$ is the set of all $n$-element permutation matrices. Thus, $|\mathcal{P}_n| = n!$ and permuting sequence $X$ by $P$ will get $X_P = \varPhi X$. We introduce some important properties of permutation matrices.
\begin{property}
    For any permutation $\varPhi \in \mathcal{P}_n$, $\varPhi^{-1} = \varPhi^\top$
\label{p:1}
\end{property}
\begin{property}
    For softmax operation, $\sigma(\varPhi X) = \varPhi\sigma(X)$ and $\sigma(X\varPhi) = \sigma(X)\varPhi $
\label{p:2}
\end{property}

Now, we explore some properties of the MHA and MLP layers.
\begin{lemma}
    For an MHA layer, input queries $Q\in\mathbb{R}^{n_q\times d_k}$, keys $K\in\mathbb{R}^{n\times d_k}$, values $V\in\mathbb{R}^{n\times d_v}$, key-value permutation $\varPhi\in\mathcal{P}_n$ and query permutation $\varPhi^Q\in\mathcal{P}_{n_q}$, we have
    \begin{equation*}
        \text{MHA}(\varPhi^Q Q, \varPhi K, \varPhi V) = \varPhi^Q \text{MHA}(Q, K, V) \text{.}
    \end{equation*}
\label{lemma:1}
\end{lemma}
\begin{proof}
    For MHA layer, the output according to Equation~\ref{eq:mha} is
    \begin{equation*}
    \begin{split}
        \text{MHA}(\varPhi^Q Q, \varPhi K, \varPhi V)\!=\!\sigma\!\left(\!\frac{\varPhi^Q Q W_i^Q {W_i^K}^\top\!K^\top\!\varPhi^\top\!}{\sqrt{d_k}}\!\right)\!\varPhi V W_i^{VO} \text{.}
    \end{split}
    \end{equation*}
    Based on Property~\ref{p:1} and~\ref{p:2}, we have
    \begin{equation*}
        \text{MHA}(\varPhi^Q Q, \varPhi K, \varPhi V) = \varPhi^Q \sigma\left(\frac{Q W_i^Q {W_i^K}^\top K^\top}{\sqrt{d_k}}\right) V W_i^{VO} \text{,}
    \end{equation*}
    which is $\varPhi^Q \text{MHA}(Q, K, V)$.
\end{proof}
\begin{property}
    For MLP operation, $\text{MLP}(\varPhi X) = \varPhi\text{MLP}(X)$.
    \label{p:3}
\end{property}

\begin{theorem}
    For a transformer encoder $E$, input sequence $X\in\mathbb{R}^{n\times d}$ and any permutation $\varPhi \in \mathcal{P}_n$, $E(\varPhi X) = \varPhi E(X)$.
\end{theorem}
\begin{proof}
    The encoder $E(X)$ can be seen as $\text{MLP} \circ \text{MHSA}(X)$ and along with Lemma~\ref{lemma:1} and Property~\ref{p:3} we have $\text{MLP} \circ \text{MHSA}(\varPhi X) = \varPhi \text{MLP} \circ \text{MHSA}(X)$.
\end{proof}

\begin{theorem}
    For a transformer decoder $D$, input memory sequence $M\in\mathbb{R}^{n\times d}$, input target sequence $T\in\mathbb{R}^{n_t\times d}$ and any memory permutation $\varPhi \in \mathcal{P}_n$, $D(\varPhi M, T) = D(M, T)$.
\end{theorem}
\begin{proof}
    The decoder of the transformer is composed of two MHA layers and one MLP layer. The first MHA layer according to~\cite{NIPS2017_attention} has the form $\text{MHA}(T, M, M)$ and consequently, $\text{MHA}(T, \varPhi M, \varPhi M) = \text{MHA}(T, M, M)$ which does not affect following layers.
\end{proof}

For the reason that the transformer is composed of stacked identical encoder and decoder layers, we have proved that the permutation $P$ applied to the input sequence $X$ will not change the final output.


\ifCLASSOPTIONcaptionsoff
    \newpage
\fi

\bibliographystyle{IEEEtran}
\bibliography{references}
\vspace{3em}

\begin{IEEEbiography}[{\includegraphics[width=1in,height=1.25in,clip,keepaspectratio]{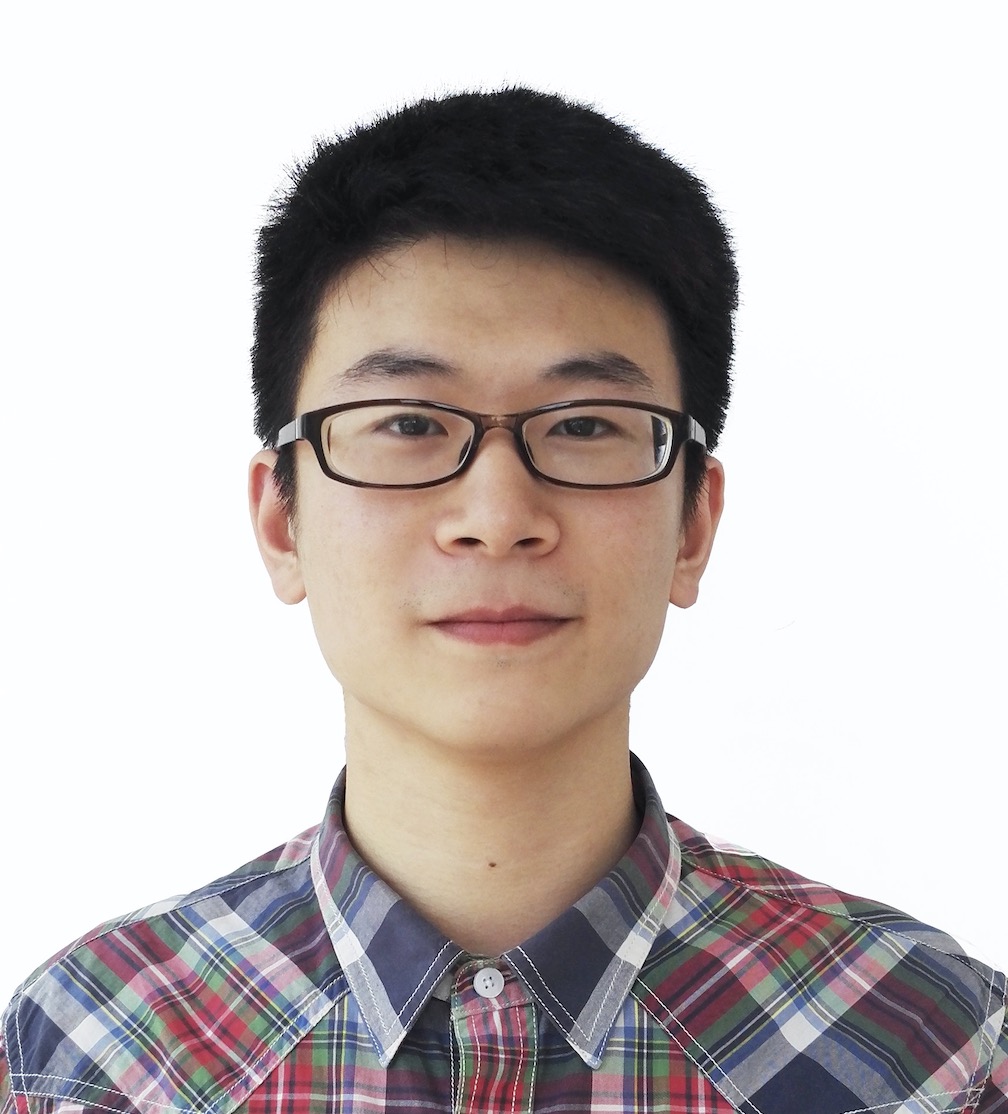}}]{Haofei~Zhang}
    is currently pursuing the Ph.D. degree with the College of Computer Science, Zhejiang University. His research interests includes transfer learning, few-shot learning, deep model reusing and interpretable machine learning.
\end{IEEEbiography}

\begin{IEEEbiography}[{\includegraphics[width=1in,height=1.25in,clip,keepaspectratio]{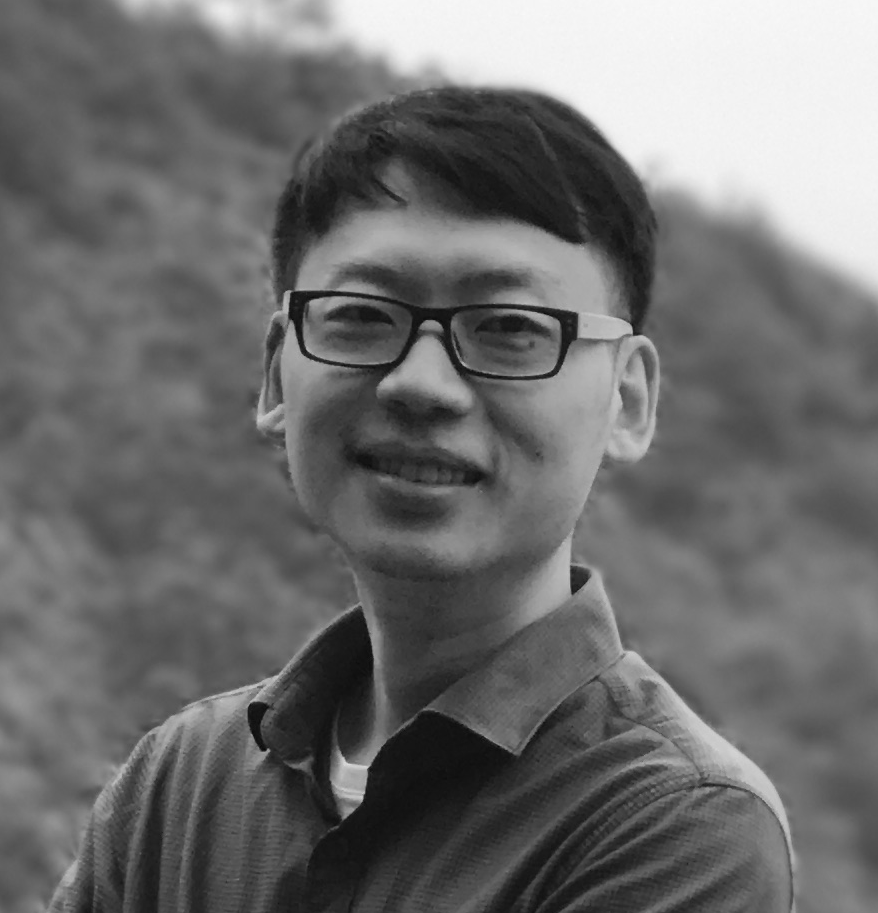}}]{Feng~Mao}
    received the master's degree from ZheJiang University in 2010. He is currently a Staff Algorithm Engineer at Alibaba Group. His current research interests include image and video understanding,  few-shot learning and multimodal learning.
\end{IEEEbiography}

\begin{IEEEbiography}[{\includegraphics[width=1in,height=1.25in,clip,keepaspectratio]{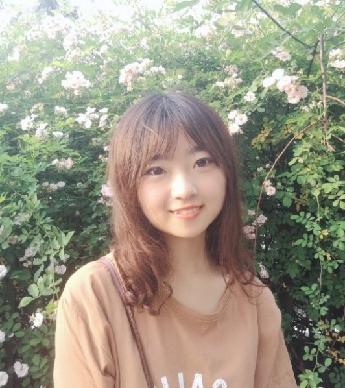}}]{Mengqi~Xue}
    is currently pursuing the Ph.D. degree with the College of Computer Science, Zhejiang University. Her research interests include multi-task learning, knowledge distillation, and vision transformers.
\end{IEEEbiography}

\begin{IEEEbiography}[{\includegraphics[width=1in,height=1.25in,clip,keepaspectratio]{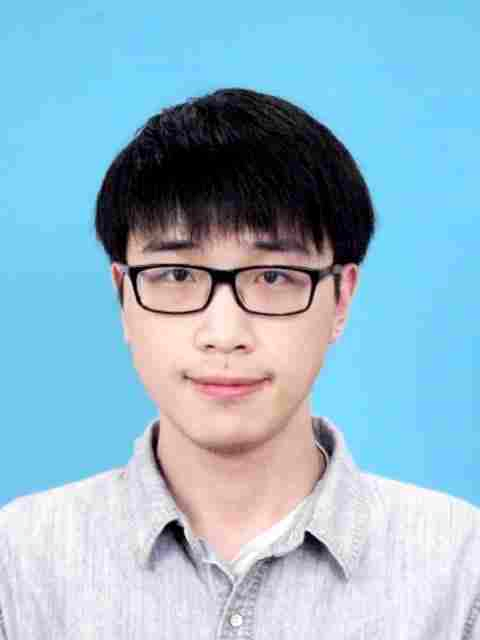}}]{Gongfan~Fang}
    is current pursuing the M.Sc. degree with the College of Computer Science, Zhejiang University. His research interests include computer vision and model compression.
\end{IEEEbiography}

\begin{IEEEbiography}[{\includegraphics[width=1in,height=1.25in,clip,keepaspectratio]{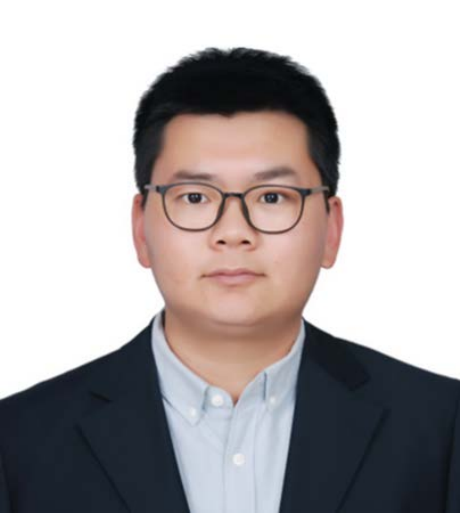}}]{Zunlei~Feng}
	is an assistant research fellow in College of Software Technology, Zhejiang University. He received his Ph.D degree in Computer Science and Technology from College of Computer Science, Zhejiang University, and B. Eng. Degree from Soochow University. His research interests mainly include computer vision, image information processing, representation learning, medical image analysis. He has authored and co-authored many scientific articles at top venues including IJCV, NeurIPS, AAAI, TVCG, ACM TOMM, and ECCV. He has served with international conferences including AAAI and PKDD, and international journals including IEEE Transactions on Circuits and Systems for Video Technology, Information Sciences, Neurocomputing, Journal of Visual Communication and Image Representation and Neural Processing Letters.
\end{IEEEbiography}

\begin{IEEEbiography}[{\includegraphics[width=1in,height=1.25in,clip,keepaspectratio]{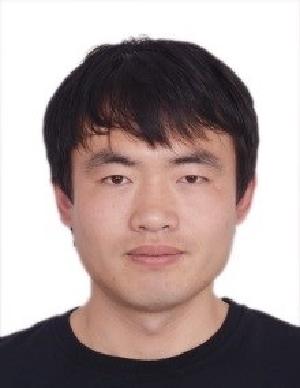}}]{Jie~Song}
	is an assistant research fellow in College of Software Technology, Zhejiang University. He received his Ph.D degree in Computer Science and Technology from College of Computer Science, Zhejiang University, and B. Eng. Degree from Sichuan University. His research interests mainly include transfer learning, few-shot learning, model compression and interpretable machine learning.
\end{IEEEbiography}

\begin{IEEEbiography}[{\includegraphics[width=1in,height=1.25in,clip,keepaspectratio]{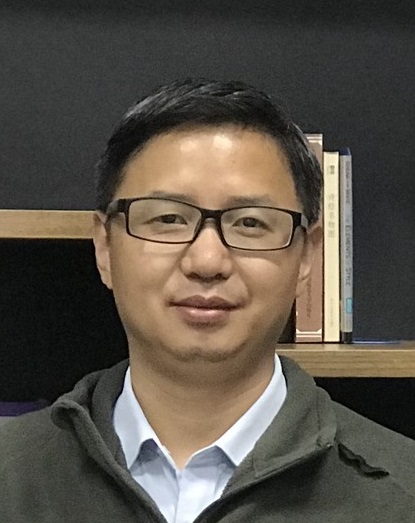}}]{Mingli~Song}
    received the Ph.D. degree in computer science from Zhejiang University, China, in 2006. He is currently a Professor with the Microsoft Visual Perception Laboratory, Zhejiang University. His research interests include face modeling and facial expression analysis. He received the Microsoft Research Fellowship in 2004.
\end{IEEEbiography}

\end{document}